\documentclass[5p,times]{elsarticle}

\usepackage{amssymb}
\usepackage{algorithmic}
\usepackage{algorithm}
\usepackage{array}
\usepackage[caption=false,font=normalsize,labelfont=sf,textfont=sf]{subfig}
\usepackage{textcomp}
\usepackage{stfloats}
\usepackage{url}
\usepackage{verbatim}
\usepackage{graphicx}
\usepackage{color}
\usepackage{booktabs}
\usepackage{xspace}
\usepackage{listings}
\usepackage{multirow}
\usepackage{colortbl}
\usepackage[table,xcdraw]{xcolor}
\usepackage{diagbox}
\usepackage{multicol}
\usepackage{lscape}
\usepackage{balance}
\usepackage{adjustbox}
\usepackage{pdflscape}
\usepackage{rotating}
\usepackage{caption}

\definecolor{green}{rgb}{0.7, 1.0, 0.7}
\definecolor{lightgreen}{rgb}{0.8, 1.0, 0.8}
\definecolor{yellow}{rgb}{1.0, 1.0, 0.6}
\definecolor{orange}{rgb}{1.0, 0.8, 0.6}

\definecolor{gray98}{rgb}{0.98,0.98,0.98}
\definecolor{gray20}{rgb}{0.20,0.20,0.20}
\definecolor{gray25}{rgb}{0.25,0.25,0.25}
\definecolor{gray16}{rgb}{0.161,0.161,0.161}
\definecolor{gray60}{rgb}{0.6,0.6,0.6}
\definecolor{gray30}{rgb}{0.3,0.3,0.3}
\definecolor{bgray}{RGB}{248, 248, 248}
\definecolor{amgreen}{RGB}{77, 175, 74}
 \definecolor{amblu}{RGB}{72, 88, 102}
\definecolor{amblu}{RGB}{55, 126, 184}
\definecolor{amred}{RGB}{180,26,28}
\definecolor{magenta}{RGB}{180,0,180}

\definecolor{lightred}{rgb}{1.0, 0.8, 0.8}
\definecolor{headerpink}{rgb}{1.0, 0.6, 0.6}
\definecolor{darkred}{rgb}{0.9, 0.4, 0.4}
\definecolor{lightgreen}{rgb}{0.8, 1.0, 0.6}
\definecolor{mediumgreen}{rgb}{0.6, 0.9, 0.4}
\definecolor{yellowgreen}{rgb}{0.9, 1.0, 0.6}
\definecolor{lightyellow}{rgb}{1.0, 1.0, 0.8}
\definecolor{lightorange}{rgb}{1.0, 0.9, 0.6}
\definecolor{darkorange}{rgb}{1.0, 0.8, 0.4}
\definecolor{yellow}{rgb}{1.0, 0.9, 0.4}

\lstdefinestyle{Ccode}{
  basicstyle=\footnotesize \tt,    
  captionpos=b,                    
  commentstyle=\color{magenta},    
  deletekeywords={},            
  escapeinside={\%*}{*)},          
  extendedchars=true,              
  frame=lines,                     
  keepspaces=true,                 
  keywordstyle=\color{blue},       
  language=C,                 
  morekeywords={},            
  numbers=none,                    
  numbersep=5pt,                   
  numberstyle=\tiny\color{gray20}, 
  rulecolor=\color{black},         
  showspaces=false,                
  showstringspaces=false,          
  showtabs=false,                  
  stepnumber=1,                    
  stringstyle=\color{mymauve},     
  tabsize=2,                       
  title=\lstname,                  
  belowskip=-1.2\baselineskip
}

\lstdefinestyle{Assemblycode}{
  basicstyle=\footnotesize \tt,    
  captionpos=b,                    
  commentstyle=\color{magenta},    
  deletekeywords={},            
  escapeinside={\%*}{*)},          
  extendedchars=true,              
  frame=lines,                     
  keepspaces=true,                 
  keywordstyle=\color{blue},       
  language=assembly,                 
  morekeywords={},            
  numbers=none,                    
  numbersep=5pt,                   
  numberstyle=\tiny\color{gray20}, 
  rulecolor=\color{black},         
  showspaces=false,                
  showstringspaces=false,          
  showtabs=false,                  
  stepnumber=1,                    
  stringstyle=\color{mymauve},     
  tabsize=2,                       
  title=\lstname,                  
  belowskip=-1.2\baselineskip
}

\newcommand{\gemm}{\textsf{GEMM}\xspace}
\newcommand{\axpy}{\textsf{AXPY}\xspace}
\newcommand{\dotp}{\textsf{DOT}\xspace}

\newcommand{\mc}{m_c}
\newcommand{\nc}{n_c}
\newcommand{\kc}{k_c}
\newcommand{\mr}{m_r}
\newcommand{\nr}{n_r}
\newcommand{\kr}{k_r}

\newcommand{\ci}{c_{i}}
\newcommand{\hi}{h_{i}}
\newcommand{\wi}{w_{i}}

\newcommand{\co}{c_{o}}
\newcommand{\ho}{h_{o}}
\newcommand{\wo}{w_{o}}

\newcommand{\hf}{h_{f}}
\newcommand{\wf}{h_{f}}

\newcommand{\lnorm}{\textsf{LNorm}\xspace}
\newcommand{\gelu}{\textsf{GELU}\xspace}
\newcommand{\conv}{\textsf{Conv}\xspace}
\newcommand{\smax}{\textsf{SoftMax}\xspace}

\journal{Future Generation Computer Systems}

\begin{document}

\begin{frontmatter}



\title{The Cambrian Explosion of Mixed-Precision Matrix Multiplication\\
       for Quantized Deep Learning Inference}

\author[label1,label2]{Héctor Martínez} 
\affiliation[label1]{organization={Universidad de Córdoba},
            country={Spain}}
            
\author[label2]{Adrián Castelló} 
\affiliation[label2]{organization={Universitat Politècnica de València},
            country={Spain}}

\author[label3]{Francisco D. Igual} 
\affiliation[label3]{organization={Universidad Complutense de Madrid},
            country={Spain}}

\author[label2]{Enrique~S.~Quintana-Ortí} 

\begin{abstract}
Recent advances in deep learning (DL) have led to a shift from traditional 64-bit floating point (FP64) computations toward reduced-precision formats—such as FP16, BF16, and 8- or 16-bit integers—combined with mixed-precision arithmetic. This transition enhances computational throughput, reduces memory and bandwidth usage, and improves energy efficiency, offering significant advantages for resource-constrained edge devices.
To support this shift, hardware architectures have evolved accordingly,
now 
including 
adapted ISAs (Instruction Set Architectures) that expose
mixed-precision vector units and matrix engines tailored for DL workloads. 

At the heart of many DL and scientific computing tasks is the general matrix-matrix multiplication (\gemm), a fundamental kernel historically optimized using \axpy vector instructions on SIMD (single instruction, multiple data) units. However, as hardware moves toward mixed-precision \dotp-product-centric operations optimized for quantized inference, these legacy approaches are being phased out.
In response to this, our paper revisits traditional high-performance \gemm and describes strategies for adapting it to mixed-precision integer (MIP) arithmetic across modern ISAs, including x86\_64, ARM, and RISC-V. Concretely, we illustrate novel micro-kernel designs and data layouts that better exploit today's specialized hardware and demonstrate significant performance gains from MIP arithmetic over floating-point implementations across three representative CPU architectures. These contributions highlight a new era of \gemm optimization-driven by the demands of DL inference on heterogeneous architectures, marking what we term as the {\em ``Cambrian period''} for matrix multiplication.
\end{abstract}




\end{frontmatter}




\section{Introduction}

In recent years, 
%
there has been a notable shift in the types of numerical representations used in computation. IEEE754 64-bit floating point precision (FP64), traditionally the backbone of scientific computing, is no longer the default in fields like 
deep learning (DL), which increasingly rely on reduced-precision formats (TensorFloat32, IEEE754 FP16, BFLOAT16, FP8, INT8, etc.) and heterogeneous (i.e., mixed-preci\-sion) arithmetic to increase throughput, lower memory demands, and reduce energy consumption while preserving 
accuracy~\cite{10.5555/3122009.3242044,10.5555/3600270.3602468,10.5555/3618408.3619993}.
Specifically, 
integer operations consume less energy and are faster than their floating-point counterparts,
particularly on specialized hardware;
and lower precision reduces memory usage and bandwidth requirements.
These factors are particularly relevant because
many DL models are deployed on edge devices with limited memory where, in addition, integer arithmetic is key to real-time performance,
latency reduction and energy efficiency~\cite{10964275}.

Accompanying this increasing heterogeneity in workloads and data formats is a parallel evolution in 
the ISA (Instruction Set Architecture) and arithmetic units. ``Classic'' SIMD (Single Instruction, Multiple Data) models 
are being replaced with mixed-precision vector units and
matrix engines (e.g., NVIDIA's Tensor Cores, Intel Advanced Matrix Extension units, Apple's Matrix Multiply Accelerator, etc.), designed specifically to accelerate a few simple linear algebra operations that are key in scientific computing and DL. These architectural changes reflect a shift toward heterogeneous domain-specific accelerators that tailor hardware to emerging computational patterns.

The general matrix-matrix multiplication (\gemm) has long been a cornerstone kernel in high-performance computing (HPC) due to its central role in scientific simulation~\cite{Kagstrom1998,demmel1997applied}.
\gemm is also crucial to the acceleration of
DL training and inference~\cite{goodfellow2016deep,vaswani2017attention,chollet2017deep},
and it is the basic building block toward the realization of 
high-performance 
BLAS (Basic Linear Algebra Subprograms) \cite{blas3}.
The efficiency of this kernel is highly sensitive to the underlying hardware, making GEMM implementations dependent on architectural changes.
For more than a decade, high-performance instances of \gemm followed a common blueprint popularized by Goto and van~de Geijn~\cite{Goto:2008:AHM:1356052.1356053}, which structured the computation around 
\axpy~\cite{blas1} operations, optimized to map efficiently onto classic SIMD units. However, this approach is increasingly being phased out as
modern hardware has shifted its focus away 
toward \textit{\dotp-centric operations that favor mixed-precision arithmetic for quantized DL inference}. This trend demands new micro-architectural strategies--such as micro-kernels that better exploit the capabilities of today’s increasingly specialized hardware,
entering what we define as the \textit{Cambrian period of mixed-preci\-sion matrix multiplication for quantized DL inference on CPU architectures}.

In this paper we review the high performance implementation of \gemm 
for quantized DL inference on multicore CPUs with a multi-level memory hierarchy and data-parallel 
arithmetic units, describing the modifications that need to be introduced in the general framework designed by
Goto and van de Geijn in order to accommodate mixed precision as well as advances in the ISA.
In doing so, we make the following contributions:
\begin{itemize}
\item We offer a complete characterization of current ISA comparing the x86\_64, RISC-V and ARMv9 
architectures.
\item We detail the implementation of the \gemm micro-kernel and associated data layout adapted to the architecture's
      ISA for ARMv8.0-A NEON, ARMv8.2 NEON, ARM SVE2, 
      specific matrix extensions for RISC-V (SpacemiT K1 integrated matrix engine, or IME),
      Intel AMX, and ARM SME.
\item We demonstrate the performance gains of MIP for computer vision and natural language
      processing inference tasks on three state-of-the-art CPUs:
      a low cost ARM Cortex-A72 CPU on a Raspberry Pi (ARMv8.0-A NEON);
      a powerful ARM Cortex-A78AE CPU on an NVIDIA Jetson AGX Orin platform (ARMv8.2); and
      a RISC-V SpacemiT K1 CPU, equipped with integer multiplication accelerators.
\end{itemize}

The rest of the paper is structured as follows.
In Section~\ref{sec:evolution}
we compile a taxonomy of current ISAs from the perspective of data-parallel arithmetic units.
In Section~\ref{sec:gemm} we revisit the original formulation of high performance \gemm
for classic SIMD units.
Next, in Sections~\ref{sec:simd}--\ref{sec:mip-mengine} we 
describe how to adapt this framework to 
\dotp-oriented SIMD units and matrix engines.
that operate on mixed precision integer (MIP) types/arithmetic.
In Section~\ref{sec:experiments} we illustrate the performance gains obtained by relying on
MIP arithmetic instead of the traditional floating point on three distinct CPUs.
Finally, we close the paper with a short summary and concluding remarks in
Section~\ref{sec:remarks}.


\section{The Evolution of Data-Parallel ISAs}
\label{sec:evolution}

\subsection{Classic SIMD ISAs}

Since the introduction of the Intel's MMX extension and its counterpart AltiVec for POWER systems by the end of the
last century, the integration of SIMD units and their corresponding ISA extensions in the micro-architecture
has become a de-facto standard in order to exploit data parallelism in virtually all processors targeting HPC workloads.

The evolution of these SIMD extensions has substantially influenced the development of optimized high-performance numerical
libraries in general, and BLAS/\gemm in particular, due to their potential and simplicity for the exploitation  of data parallelism.
In consequence, the release of new optimized versions of BLAS, both proprietary (e.g., MKL for Intel, ESSL for POWER,
AOCL for AMD, ARMPL for ARM, \ldots) and academic (GotoBLAS, BLIS, OpenBLAS, \ldots),
has mainly evolved hand in hand with the micro-architectural and ISA evolution.
Table~\ref{tab:isas} summarizes the last 25 years of ISA extensions, 
highlighting the type of instructions that are typically involved in the processing of the arithmetic
workload within BLAS,
%
for the three main actors in
the current HPC arena: x86\_64, ARM and RISC-V.
Beyond the description and characterization of each instruction, the table also reports empirical performance
numbers (in terms of GOPS --billions of arithmetic operations per second--) observed on four state-of-the-art platforms
that implement these ISAs, and the relative performance compared with the corresponding
FMA (Fused Mul\-tiply-Accu\-mulate) instruction operating on FP32 data, 
typically used in {\em classic} BLAS implementations relying in the original approach by Goto 
and van de Geijn.

Taking x86\_64 ISA as a driving example, the evolution from SSE
({\em Streaming SIMD Extensions}, released in 1999) to
AVX and AVX2 (Advanced Vector Extensions, released in 2011)
up to AVX-512 (released in 2016) shows illustrates a trend that %
%
impacted the design of
BLAS from two main perspectives: 
\begin{enumerate}
\item Increasing the vector width and number of vector registers (from 16 128-bit registers in SSE to 32 512-bit registers in AVX512).
\item Adding functionality to increase performance and numerical stability 
for the \axpy operations (e.g. the introduction of FMA instructions in AVX2 and AVX-512).
\end{enumerate}
More recently the ISAs were modified to also accommodate (homogeneous) reduced-precision 
arithmetic (e.g. FP16). This feature is now present in x86\_64 as well as modern ARM and RISC-V 
vector extensions.
All in all, until the last decade, the support of FMA-oriented instructions and single-type (i.e.,
homogeneous) SIMD arithmetic units 
gave the necessary support to BLAS developers to exploit the underlying 
vector capabilities by means of \axpy-based micro-kernels (see Section~\ref{sec:axpy}) that 
supported the whole BLAS functionality necessary for scientific computing.

\subsection{\dotp-oriented SIMD ISAs and mixed precision}

By the end of the last decade, and mainly in response to the requirements of DL workloads, 
modern ISAs have evolved in two intertwined directions.
First, towards the introduction of \dotp-oriented instructions.
Second, towards the integration of mixed precision in all
\dotp-oriented instructions.
Continuing with the x86\_64 example, the BF16 and VNNI extensions to the base AVX-512 ISA introduced support for BF16 and integer (INT8, INT16) input operands in the \dotp instruction, with accumulation into 32-bit floating-point (FP32) and integer (INT32), respectively.
As Table~\ref{tab:isas} shows, in all cases the performance acceleration derived from the use of these
instructions is 
remarkable\footnote{The use of reduced precision also encompasses less memory and bandwidth pressure, even though this improvement is not reported in the table.}.


\begin{landscape}
\begin{table}
\centering
\scriptsize
\begin{tabular}{lllllllllllllccc}
\multicolumn{1}{c}{}                                                                              & \multicolumn{1}{c}{}                                                                                   & \multicolumn{1}{c}{}                                & \multicolumn{1}{c}{}                                       & \multicolumn{1}{c}{}                                                                               &                                      & \multicolumn{1}{c}{}                                                                          & \multicolumn{1}{c}{}                                                                                   & \multicolumn{1}{c}{}                                                                                   & \multicolumn{1}{c}{}                                                                                     & \multicolumn{1}{c}{}                                                                                             & \multicolumn{1}{c}{}                                                                                       & \multicolumn{1}{c}{}                                                                               &                                                                                                                    &                                                                                          &                                                                                           \\
\multicolumn{1}{c}{}                                                                              & \multicolumn{1}{c}{}                                                                                   & \multicolumn{1}{c}{}                                & \multicolumn{1}{c}{}                                       & \multicolumn{1}{c}{}                                                                               &                                      & \multicolumn{1}{c}{}                                                                          & \multicolumn{1}{c}{}                                                                                   & \multicolumn{1}{c}{}                                                                                   & \multicolumn{1}{c}{}                                                                                     & \multicolumn{1}{c}{}                                                                                             & \multicolumn{1}{c}{}                                                                                       & \multicolumn{1}{c}{}                                                                               &                                                                                                                    &                                                                                          &                                                                                           \\
\multicolumn{1}{c}{\multirow{-3}{*}{\textbf{\begin{tabular}[c]{@{}c@{}}Base\\ ISA\end{tabular}}}} & \multicolumn{1}{c}{\multirow{-3}{*}{\textbf{\begin{tabular}[c]{@{}c@{}}ISA\\ Extension\end{tabular}}}} & \multicolumn{1}{c}{\multirow{-3}{*}{\textbf{Year}}} & \multicolumn{1}{c}{\multirow{-3}{*}{\textbf{Instruction}}} & \multicolumn{1}{c}{\multirow{-3}{*}{\textbf{\begin{tabular}[c]{@{}c@{}}ISA \\ Type\end{tabular}}}} & \multirow{-3}{*}{\textbf{Precision}} & \multicolumn{1}{c}{\multirow{-3}{*}{\textbf{State}}}                                          & \multicolumn{1}{c}{\multirow{-3}{*}{\textbf{\begin{tabular}[c]{@{}c@{}}Example\\ arch.\end{tabular}}}} & \multicolumn{1}{c}{\multirow{-3}{*}{\textbf{\begin{tabular}[c]{@{}c@{}}Matrix\\ engine\end{tabular}}}} & \multicolumn{1}{c}{\multirow{-3}{*}{\textbf{\begin{tabular}[c]{@{}c@{}}\# \\ Tile\\ regs.\end{tabular}}}} & \multicolumn{1}{c}{\multirow{-3}{*}{\textbf{\begin{tabular}[c]{@{}c@{}}Tile reg.\\ dim.\\ (bits)\end{tabular}}}} & \multicolumn{1}{c}{\multirow{-3}{*}{\textbf{\begin{tabular}[c]{@{}c@{}}\# \\ Vector\\ regs.\end{tabular}}}} & \multicolumn{1}{c}{\multirow{-3}{*}{\textbf{\begin{tabular}[c]{@{}c@{}}VL\\ (bits)\end{tabular}}}} & \multirow{-3}{*}{\textbf{\begin{tabular}[c]{@{}c@{}}Operand\\ dimensions\\ ($m \times n  \times k$)\end{tabular}}} & \multirow{-3}{*}{\textbf{\begin{tabular}[c]{@{}c@{}}Peak\\ Perf.\\ (GOPS)\end{tabular}}} & \multirow{-3}{*}{\textbf{\begin{tabular}[c]{@{}c@{}}Ratio\\ vs. FP32\\ FMA\end{tabular}}} \\ \hline
                                                                                                  & AMX\_INT8                                                                                              & 2023                                                & tdpbssd(i32,i8,i8)                                         & Matrix                                                                                             & Mixed                                &                                                                                               &                                                                                                        & \cellcolor[HTML]{FFFFFF}                                                                               & 8                                                                                                        & 1024                                                                                                             & -                                                                                                          & -                                                                                                  & $16 \times 16 \times 64$                                                                                           & 7776.40                                                                                  & 65.36                                                                                     \\
                                                                                                  & AMX\_BF16                                                                                              & 2023                                                & tdpbf16ps(f32,bf16,bf16)                                   & Matrix                                                                                             & Mixed                                & \multirow{-2}{*}{\begin{tabular}[c]{@{}l@{}}Separated,\\ Tile-oriented\end{tabular}}          &                                                                                                        & \multirow{-2}{*}{\cellcolor[HTML]{FFFFFF}Yes}                                                          & 8                                                                                                        & 1024                                                                                                             & -                                                                                                          & -                                                                                                  & $16 \times 16 \times 32$                                                                                           & 3900.80                                                                                  & 33.09                                                                                     \\ \cline{2-7} \cline{9-16} 
                                                                                                  & AVX\_VNNI\_INT8                                                                                        & 2021                                                & vpdpbssd(i32,i8,i8)                                        & DotProd                                                                                            & Mixed                                & -                                                                                             &                                                                                                        & -                                                                                                      & -                                                                                                        & -                                                                                                                & 16                                                                                                         & 256                                                                                                & $16 \times 16 \times 4$                                                                                            & 908.87                                                                                   & 7.71                                                                                      \\
                                                                                                  & AVX512\_VNNI                                                                                           & 2021                                                & vpdpwssd(i32,i16,i16)                                      & DotProd                                                                                            & Mixed                                & -                                                                                             &                                                                                                        & -                                                                                                      & -                                                                                                        & -                                                                                                                & 32                                                                                                         & 512                                                                                                & $16 \times 16 \times 2$                                                                                            & 482.10                                                                                   & 4.09                                                                                      \\
                                                                                                  & AVX512\_BF16                                                                                           & 2020                                                & vdpbf16ps(f32,bf16,bf16)                                   & DotProd                                                                                            & Mixed                                & -                                                                                             &                                                                                                        & -                                                                                                      & -                                                                                                        & -                                                                                                                & 32                                                                                                         & 512                                                                                                & $16 \times 16 \times 32$                                                                                           & 118.53                                                                                   & 1.01                                                                                      \\ \cline{2-7} \cline{9-16} 
                                                                                                  & AVX512\_FP16                                                                                           & 2019                                                & vfmadd231ph(f16,f16,f16)                                   & FMA                                                                                                & HP                                   & -                                                                                             &                                                                                                        & -                                                                                                      & -                                                                                                        & -                                                                                                                & 32                                                                                                         & 512                                                                                                & $32 \times 1 \times 32$                                                                                            & 466.59                                                                                   & 3.96                                                                                      \\
                                                                                                  & AVX512F                                                                                                & 2016                                                & vfmadd231ps(f32,f32,f32)                                   & FMA                                                                                                & SP                                   & -                                                                                             &                                                                                                        & -                                                                                                      & -                                                                                                        & -                                                                                                                & 32                                                                                                         & 512                                                                                                & $16 \times 1 \times 16$                                                                                            & 117.89                                                                                   & 1.00                                                                                      \\
                                                                                                  & AVX512F                                                                                                & 2016                                                & vfmadd231pd(f64,f64,f64)                                   & FMA                                                                                                & DP                                   & -                                                                                             &                                                                                                        & -                                                                                                      & -                                                                                                        & -                                                                                                                & 32                                                                                                         & 512                                                                                                & $8 \times 1 \times 8$                                                                                              & 60.38                                                                                    & 0.50                                                                                      \\ \cline{2-7} \cline{9-16} 
                                                                                                  & AVX2                                                                                                   & 2011                                                & vaddps(vmulps(f32,f32,f32))                                & ADD+MUL                                                                                            & SP                                   & -                                                                                             &                                                                                                        & -                                                                                                      & -                                                                                                        & -                                                                                                                & 16                                                                                                         & 256                                                                                                & $8 \times 1 \times 8$                                                                                              & 86.72                                                                                    & 0.74                                                                                      \\
                                                                                                  & AVX2                                                                                                   & 2011                                                & vaddps(vmulps(f64,f64,f64))                                & ADD+MUL                                                                                            & DP                                   & -                                                                                             &                                                                                                        & -                                                                                                      & -                                                                                                        & -                                                                                                                & 16                                                                                                         & 256                                                                                                & $4 \times 1 \times 4$                                                                                              & 43.42                                                                                    & 0.37                                                                                      \\
                                                                                                  & SSE4                                                                                                   & 2000                                                & addps(mulps(f32,f32,f32))                                  & ADD+MUL                                                                                            & SP                                   & -                                                                                             &                                                                                                        & -                                                                                                      & -                                                                                                        & -                                                                                                                & 16                                                                                                         & 128                                                                                                & $4 \times 1 \times 4$                                                                                              & 46.32                                                                                    & 0.39                                                                                      \\
\multirow{-12}{*}{\textbf{x86\_64}}                                                               & SSE4                                                                                                   & 2000                                                & addps(mulps(f64,f64,f64))                                  & ADD+MUL                                                                                            & DP                                   & -                                                                                             & \multirow{-12}{*}{\begin{tabular}[c]{@{}l@{}}Sapphire\\ Rapids\end{tabular}}                           & -                                                                                                      & -                                                                                                        & -                                                                                                                & 16                                                                                                         & 128                                                                                                & $2 \times 1 \times 2$                                                                                              & 22.52                                                                                    & 0.19                                                                                      \\ \hline
                                                                                                  & IME                                                                                                    & 2024                                                & vmadot(i32,i8,i8)                                          & Matrix                                                                                             & Mixed                                &                                                                                               &                                                                                                        &                                                                                                        & -                                                                                                        & -                                                                                                                & 32                                                                                                         & 256                                                                                                & $4 \times 4 \times 8$                                                                                              & 406.36                                                                                   & 15.33                                                                                     \\
                                                                                                  & IME                                                                                                    & 2024                                                & vfmadot(f4-16,f4-16,f4-16)                                 & Matrix                                                                                             &                                      & \multirow{-2}{*}{\begin{tabular}[c]{@{}l@{}}Common,\\ vector-oriented\end{tabular}}           &                                                                                                        & \multirow{-2}{*}{No}                                                                                   & -                                                                                                        & -                                                                                                                & 32                                                                                                         & 256                                                                                                &                                                                                                                    & -                                                                                        & -                                                                                         \\ \cline{2-7} \cline{9-16} 
                                                                                                  & RVV-1.0                                                                                                & 2021                                                & vfmacc.vf/vv(f16,f16,f16)                                  & FMA                                                                                                & HP                                   & -                                                                                             &                                                                                                        & -                                                                                                      & -                                                                                                        & -                                                                                                                & 32                                                                                                         & 256                                                                                                & $32 \times 1 \times 32$                                                                                            & 52.99                                                                                    & 2.00                                                                                      \\
                                                                                                  & RVV-1.0                                                                                                & 2021                                                & vfmacc.vf/vv(f32,f32,f32)                                  & FMA                                                                                                & SP                                   & -                                                                                             &                                                                                                        & -                                                                                                      & -                                                                                                        & -                                                                                                                & 32                                                                                                         & 256                                                                                                & $16 \times 1 \times 16$                                                                                            & 26.50                                                                                    & 1.00                                                                                      \\
\multirow{-5}{*}{\textbf{RISC-V}}                                                                 & RVV-1.0                                                                                                & 2021                                                & vfmacc.vf/vv(f64,f64,f64)                                  & FMA                                                                                                & DP                                   & -                                                                                             & \multirow{-5}{*}{\begin{tabular}[c]{@{}l@{}}SpacemiT \\ K1\end{tabular}}                               & -                                                                                                      & -                                                                                                        & -                                                                                                                & 32                                                                                                         & 256                                                                                                & $8 \times 1 \times 8$                                                                                              & 13.25                                                                                    & 0.50                                                                                      \\ \hline
                                                                                                  & SME2                                                                                                   & 2022                                                & fmopa(f32,f32,f32)                                         & Matrix                                                                                             & SP                                   &                                                                                               &                                                                                                        &                                                                                                        & 1 (4)                                                                                                    & 4096                                                                                                             & 32                                                                                                         & 512                                                                                                & $16 \times 16 \times 1$                                                                                            & 2004.36                                                                                  & 16.95                                                                                     \\
                                                                                                  & SME2+F64\_F64                                                                                          & 2022                                                & fmopa(f64,f64,f64)                                         & Matrix                                                                                             & DP                                   &                                                                                               &                                                                                                        &                                                                                                        & 1 (8)                                                                                                    & 4096                                                                                                             & 32                                                                                                         & 512                                                                                                & $8 \times 8 \times 1$                                                                                              & 501.31                                                                                   & 4.24                                                                                      \\
                                                                                                  & SME2+F16\_F16                                                                                          & 2022                                                & fmopa(f32,f16,f16)                                         & Matrix                                                                                             & Mixed                                &                                                                                               &                                                                                                        &                                                                                                        & 1 (2)                                                                                                    & 4096                                                                                                             & 32                                                                                                         & 512                                                                                                & $16 \times 16 \times 2$                                                                                            & 2006.34                                                                                  & 16.97                                                                                     \\
                                                                                                  & SME2+B16\_B16                                                                                          & 2022                                                & bfmopa(f32,bf16,bf16)                                      & Matrix                                                                                             & Mixed                                &                                                                                               &                                                                                                        &                                                                                                        & 1 (2)                                                                                                    & 4096                                                                                                             & 32                                                                                                         & 512                                                                                                & $16 \times 16 \times 2$                                                                                            & 2006.23                                                                                  & 16.97                                                                                     \\
                                                                                                  & SME2                                                                                                   & 2022                                                & smopa(i32,i16,i16)                                         & Matrix                                                                                             & Mixed                                &                                                                                               &                                                                                                        &                                                                                                        & 1 (2)                                                                                                    & 4096                                                                                                             & 32                                                                                                         & 512                                                                                                & $16 \times 32 \times 1$                                                                                            & 2006.71                                                                                  & 16.97                                                                                     \\
                                                                                                  & SME2                                                                                                   & 2022                                                & smopa(i32,i8,i8)                                           & Matrix                                                                                             & Mixed                                &                                                                                               &                                                                                                        &                                                                                                        & 1 (4)                                                                                                    & 4096                                                                                                             & 32                                                                                                         & 512                                                                                                & $16 \times 16 \times 4$                                                                                            & 4013.41                                                                                  & 33.94                                                                                     \\
\multirow{-7}{*}{\textbf{ARMv9}}                                                                  & SME2                                                                                                   & 2022                                                & smopa(i64,i16,i16)                                         & Matrix                                                                                             & Mixed                                & \multirow{-7}{*}{\begin{tabular}[c]{@{}l@{}}Separated,\\ vector/tile\\ oriented\end{tabular}} & \multirow{-7}{*}{\begin{tabular}[c]{@{}l@{}}Apple \\ M4\end{tabular}}                                  & \multirow{-7}{*}{Yes}                                                                                  & 1 (8)                                                                                                    & 4096                                                                                                             & 32                                                                                                         & 512                                                                                                & $8 \times 8 \times 4$                                                                                              & 2006.21                                                                                  & 16.97                                                                                     \\ \hline
                                                                                                  & SVE2+I8MM                                                                                              & 2019                                                & mmla(i32,i8,i8)                                            & Matrix                                                                                             & Mixed                                &                                                                                               &                                                                                                        &                                                                                                        & -                                                                                                        & -                                                                                                                & 32                                                                                                         & 128                                                                                                & $2 \times 2 \times 8$                                                                                              & 834.09                                                                                   & 8.00                                                                                      \\
                                                                                                  & SVE2+BF16                                                                                              & 2019                                                & bfmmla(f32,bf16,bf16)                                      & Matrix                                                                                             & Mixed                                & \multirow{-2}{*}{\begin{tabular}[c]{@{}l@{}}Common,\\ vector-oriented\end{tabular}}           &                                                                                                        & \multirow{-2}{*}{No}                                                                                   & -                                                                                                        & -                                                                                                                & 32                                                                                                         & 128                                                                                                & $2 \times 2 \times 4$                                                                                              & 386.98                                                                                   & 3.71                                                                                      \\ \cline{2-7} \cline{9-16} 
                                                                                                  & SVE2+DotProd                                                                                           & 2019                                                & sdot.vs/vv(i32,i8,i8)                                      & DotProd                                                                                            & Mixed                                & -                                                                                             &                                                                                                        & -                                                                                                      & -                                                                                                        & -                                                                                                                & 32                                                                                                         & 128                                                                                                & $4 \times 4 \times 4$                                                                                              & 417                                                                                      & 4.00                                                                                      \\
                                                                                                  & SVE2+BF16                                                                                              & 2019                                                & bfdot.vs/vv(f32,bf16,bf16)                                 & DotProd                                                                                            & Mixed                                & -                                                                                             &                                                                                                        & -                                                                                                      & -                                                                                                        & -                                                                                                                & 32                                                                                                         & 128                                                                                                & $4 \times 4 \times 2$                                                                                              & 417.04                                                                                   & 2.00                                                                                      \\ \cline{2-7} \cline{9-16} 
                                                                                                  & SVE2+FP16                                                                                              & 2016                                                & fmla.vs/vv(f16,f16,f16)                                    & FMA                                                                                                & HP                                   & -                                                                                             &                                                                                                        & -                                                                                                      & -                                                                                                        & -                                                                                                                & 32                                                                                                         & 128                                                                                                & $8 \times 1 \times 8$                                                                                              & 208.55                                                                                   & 2.00                                                                                      \\
                                                                                                  & SVE2                                                                                                   & 2011                                                & fmla.vs/vv(f32,f32,f32)                                    & FMA                                                                                                & SP                                   & -                                                                                             &                                                                                                        & -                                                                                                      & -                                                                                                        & -                                                                                                                & 32                                                                                                         & 128                                                                                                & $4 \times 1 \times 4$                                                                                              & 104.29                                                                                   & 1.00                                                                                      \\
\multirow{-7}{*}{\textbf{ARMv9}}                                                                  & SVE2                                                                                                   & 2011                                                & fmla.vs/vv(f64,f64,f64)                                    & FMA                                                                                                & DP                                   & -                                                                                             & \multirow{-7}{*}{\begin{tabular}[c]{@{}l@{}}NVIDIA  \\ Grace \\ CPU\end{tabular}}                       & -                                                                                                      & -                                                                                                        & -                                                                                                                & 32                                                                                                         & 128                                                                                                & $2 \times 1 \times 2$                                                                                              & 52.14                                                                                    & 0.50                                                                                      \\ \hline
\end{tabular}
\caption{Overview of the evolution of ISAs and instructions related to \gemm micro-kernel development.
DP=FP64, SP=FP32, HP=FP16, VL=Vector Length.}
\label{tab:isas}
\end{table}
\end{landscape}

\subsection{Tile-oriented ISAs, matrix engines and mixed precision}

The design of tile-oriented ISA extensions was explored at the end of the last decade
in response to demands of DL workloads, and was premiered for CPUs by Intel through the AMX ISA extension~\cite{KimLLMAMX,abouelhamayed2025sparamxacceleratingcompressedllms} 
(introduced in 2020 and implemented for the first time in the Sapphire Rapids architecture, back in 2023). 
This idea extended the novelty of the I8MM  extension~\cite{xiao2025understandinglargelanguagemodels} 
to the base ARMv8 ISA in 2019, and consolidated the use of bi-dimensional operands as source and destination 
of the corresponding SIMD instructions. 
A similar approach was recently proposed by the proprietary matrix extension for RISC-V in the SpacemiT K1~\cite{githubReleasesSpacemitriscvimeextensionspec}, 
which mimics some of the ideas that are under development for the {\em standardized} IME 
and AME (Attached Matrix Extension) to integrate them as extensions to RISC-V. 
Finally, the Scalable Matrix Extension (SME) builds upon SVE as ARM's response to advancements in other ISAs, and is increasingly adopted in recent processors—such as the Apple M4~\cite{remke2024hellosmegeneratingfast}—to provide tile-oriented functionality.

Beyond accommodating the matrix abstraction in low-level kernel development, which heavily increases programmability, 
the use of tile-oriented instructions also  enables the adoption of {\em matrix engines} to improve performance
through specialized accelerators. 
These extensions are an atractive way of exposing tightly-coupled matrix computation accelerators
via simple ISA extensions, with excellent support from compilers (even via intrinsics).
Actually, the design of matrix engines was already pioneered by NVIDIA
starting with the Volta microarchitecture, but tile-oriented ISAs democratizes their use in CPU architectures. 
The TMUL~\cite{Intel_2024} ({\em Tile Matrix Multiply}) unit from Intel, usable via the AMX ISA extension,  comprises a grid of FMA 
units that can operate on tiles, and is able to  perform matrix multiplication on INT8 or BF16 inputs. Similarly, Apple's
Matrix Multiply Accelerator, integrated in the latest M4 processor, is for the first time programmable via ARM SME, which 
opens a research avenue to port existing software to this new matrix engine, avoiding the use 
of proprietary libraries and
frameworks.

\section{\gemm Macro-kernel}
\label{sec:gemm}

For over 15 years, high-performance implementations of \gemm in scientific computing libraries--such as AMD AOCL, IBM ESSL, ARM PL, GotoBLAS2, OpenBLAS, BLIS, and presumably Intel MKL--have largely adhered to the principles introduced by Goto and van de Geijn in~\cite{Goto:2008:AHM:1356052.1356053}. The present section and the following one review this framework
in the context of CPUs equipped with deep memory hierarchies and \textit{classic, \axpy-oriented SIMD units that process single-type data.}

The reference \gemm algorithm--depicted in Figure~\ref{fig:baseline_GEMM} and also known as the \textit{macro-kernel}--is composed of five nested loops, labeled {\tt L1} through {\tt L5}. To enhance memory access efficiency and cache utilization, it includes two packing stages that reorganize certain blocks of $A$ and $B$ into two contiguous buffers, $A_c$ and $B_c$ respectively. At the core of the computation lies a \textit{micro-kernel}, responsible for performing the multiply-accumulate operations within a sixth loop discussed in the next section.

\begin{figure}[tbh]
\centering
\begin{minipage}[t]{0.9\columnwidth}
\begin{tabular}{l}
\begin{lstlisting}[language=C,alsoletter={.},deletekeywords={.sum},morekeywords={float32x4_t}]
for (jc=0; jc<n; jc+=nc)                // Loop L1
  for (pc=0; pc<k; pc+=kc) {            // Loop L2
    Bc := B(pc:pc+kc-1,jc:jc+nc-1);     // Pack B
    for (ic=0; ic<m; ic+=mc) {          // Loop L3
      Ac := A(ic:ic+mc-1,pc:pc+kc-1);   // Pack A
      for (jr=0; jr<nc; jr+=nr)         // Loop L4
        for (ir=0; ir<mc; ir+=mr)       // Loop L5
          // Micro-kernel
          C(ic+ir:ic+ir+mr-1, jc+jr:jc+jr+nr-1)  
                 += Ac(ir:ir+mr-1,0:kc-1) 
                  * Bc(0:kc-1,jr:jr+nr-1);
  } } 
\end{lstlisting}
\end{tabular}
\end{minipage}
\caption{Reference \gemm algorithm (macro-kernel) in GotoBLAS2.}
\label{fig:baseline_GEMM}
\end{figure}

To fully optimize \gemm on modern multicore CPUs, developers must address three key challenges:
\begin{itemize} 
\item Carefully tune the blocking (tiling) parameters $\mc$, $\nc$, $\kc$ to make effective use of the cache hierarchy~\cite{Goto:2008:AHM:1356052.1356053,BLIS4}. 
\item Ensure ample parallelism within the loop structure to facilitate a balanced workload distribution across CPU cores and to leverage both private and shared cache levels efficiently~\cite{BLIS2,BLIS3}. 
\item Design a high-performance micro-kernel and associated packing routines to maximize arithmetic throughput. 
\end{itemize}

To close this high level discussion of \gemm, we note that
a general implementation must support transposed inputs $A$ and/or $B$, as well as different storage orders (row-major or column-major) for all three matrix operands. The effects of transposition and storage layout for $A$ and $B$ can be encapsulated within their respective packing routines, keeping the micro-kernel agnostic to these details. 
In addition, a column-major layout for $C$ can be handled by swapping the roles of $A$ and $B$ when invoking the 
micro-kernel.

In the following sections, we examine the evolution of the micro-kernel and the corresponding packing routines from their original formulation in~\cite{Goto:2008:AHM:1356052.1356053} for FP32/FP64 arithmetic on classic SIMD-enabled CPUs (Section~\ref{sec:simd}), to a heterogeneous collection of solutions 
for MIP arithmetic on specialized units (Sections~\ref{sec:mip-simd} and~\ref{sec:mip-mengine}). 
In all these
cases, the macro-kernel remains basically the same 
presented in Figure~\ref{fig:baseline_GEMM}.

\section{Micro-kernel for Classic, \axpy-oriented SIMD Units}
\label{sec:simd}

SIMD units are specialized hardware components designed to execute the same operation on multiple data simultaneously, significantly boosting performance for data-parallel wokloads.
The throughput of SIMD units is closely tied to the width of the data paths and the number of parallel lanes, which determine how many operations can be completed per instruction cycle. For instance, the 512-bit SIMD unit in Intel AVX-512 can operate on sixteen FP32 numbers simultaneously, and this performance scaling is further enhanced by the hardware support for FMA, which combines a multiplication and an addition in a single instruction. The availability and implementation of such operations are dictated by the processor's ISA, which defines the set of SIMD instructions, register widths, supported data types, and the semantics of operations like FMA. 

GotoBLAS2 was developed during the era of x86's SSE4 SIMD extension (vector length VL=128 bits), 
and adapted effectively to subsequent ISA advancements:
AVX, AVX2 (both with VL=256 bits), 
and 
AVX-512 (VL=512 bits). 
Its architecture also aligned well with the evolution of ARM designs, including NEON (VL=128 bit) 
and SVE (variable VL), 
and more recently with RISC-V Vector (RVV) 1.0 (also variable VL). 

In this section we review how GotoBLAS2 originally formulated a micro-kernel together with an
accompanying data packing schemes 
in order to exploit the scalability of FMA instructions on
classic \axpy-oriented SIMD units that operate on
single-type
(i.e., homogeneous) floating-point data.

\subsection{Overview of the micro-kernel}
Consider the update of $C$ in lines 8--11 of the reference algorithm in Figure~\ref{fig:baseline_GEMM}, 
corresponding to the \textit{micro-kernel}. This component performs a small \gemm of the form $C_r \,+\!\!= A_rB_r$, where:
\begin{itemize}
\setlength{\itemsep}{0pt}
\item $C_r$ is an $\mr \times \nr$ micro-tile (or block) of matrix $C$,
\item $A_r$ is an $\mr \times \kc$ micro-panel (or slab) of buffer $A_c$, and
\item $B_r$ is a $\kc \times \nc$ micro-panel of buffer $B_c$.
\end{itemize}
GotoBLAS2 implements the micro-kernel through an additional loop, labeled as~\texttt{L6}, that iterates over the $k_c$ dimension of the problem, at each iteration performing a rank-1 update of $C_r$ involving a column of $A_r$ and a row of $B_r$.
In addition, the micro-kernel loads $C_r$ into vector registers before entering loop~\texttt{L6}, and writes 
its contents back to memory after the loop completes. 
For large $\kc$, this design amortizes the memory access cost for $C_r$ across sufficient arithmetic operations.
Furthermore, the specialized packing of $A_c$ and $B_c$ into micro-panels of $m_r$ rows and $n_r$ columns respectively, as illustrated in the top part of Figure~\ref{fig:gotoblas2_micro}, ensures that their elements can be accessed with stride-1 during the execution of the
micro-kernel loop body. Concretely,
the white arrows in the packing scheme specify the layout of the elements of $A_c,B_c$ in memory.
This organization is well-suited for efficient use of vector load instructions.
Finally, the overhead of packing the data is amortized by reusing the contents of $A_c,B_c$ multiple times
(e.g., the elements of $A_c$ are accessed across the iterations of
loops~\texttt{L4} and \texttt{L5}, that is, $\lceil\mc/\mr\rceil \lceil\nc/\nr\rceil$ times).

\subsection{Choosing $\mr, \nr$}
The arithmetic intensity of an algorithm
is defined as the ratio of arithmetic operations to memory traffic, and serves
as an indicator of whether the algorithm is memory-bound or compute-bound~\cite{williams2009roofline}. 
In the case of the micro-kernel, this component conducts $2\mr\nr\kc$ ``useful'' arithmetic operations, while accessing
$\mr\nr$ elements (read and write) for $C_r$;
$\mr\kc$ elements (read) for $A_r$; and
$\kc\nr$ elements (read) for $B_r$.
Since $\kc$ is typically set based on parameters of the \textsf{L1} cache~\cite{BLIS4}, the optimal strategy to maximize 
the arithmetic intensity is to choose $\mr \approx \nr$ and make both parameters as large as possible. However, a practical limit arises: to avoid register spilling, the vector register file must accommodate the entire micro-tile $C_r$, plus at least a column of $A_r$ and a row of $B_r$. This constraint imposes an upper bound on $\mr, \nr$.

\subsection{Matching the micro-kernel to \axpy-oriented SIMD units\label{sec:axpy}}

Assume hereafter that matrix $C$ is stored in row-major order 
with leading dimension \texttt{ldC},
and let us consider the rank-1 update occurring at iteration $k$ of loop \texttt{L6}: $C_r \,+\!\!= a_rb_r$, where $a_r$ and $b_r$ respectively denote the $k$-th column and row of  the micro-panels $A_r$ and $B_r$.
To align with classic SIMD units, the micro-kernel decomposes this rank-1 update into several \axpy operations, each updating a portion of a row of $C_r$ by multiplying one element of $a_r$ with part of a row of $b_r$.

\begin{figure*}[t!]
\begin{center}
\includegraphics[width=0.7\textwidth]{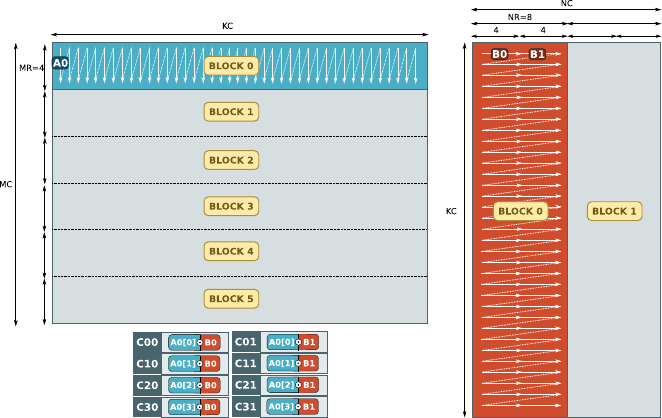}
\centering
\begin{minipage}{0.9\linewidth}
\begin{lstlisting}[linewidth=\textwidth, language=C,alsoletter={.},deletekeywords={.sum},morekeywords={v4_fp32,float32x4_t},breaklines=true]
#define Cr(i, j) C[(i)*ldC+(j)]
float32x4_t A0, B0, B1, C00, C01, C10, C11, C20, C21, C30, C31;

// Load initial Cr
vld1q_f32(&Cr(0,0), C00); vld1q_f32(&Cr(0,4), C01); vld1q_f32(&Cr(1,0), C10); vld1q_f32(&Cr(1,4), C11); 
vld1q_f32(&Cr(2,0), C20); vld1q_f32(&Cr(2,4), C21); vld1q_f32(&Cr(3,0), C30); vld1q_f32(&Cr(3,4), C31);
// Micro-kernel
for (int k=0; k<kc; k++) { // Loop L6
  vld1q_f32(&Ar[baseA], A0); vld1q_f32(&Br[baseB], B0); vld1q_f32(&Br[baseB+4], B1); 
  baseA+=4; baseB+=8;
  C00=vfmaq_laneq_f32(C00, B0, A0, 0); C01=vfmaq_laneq_f32(C01, B1, A0, 0); 
  C10=vfmaq_laneq_f32(C10, B0, A0, 1); C11=vfmaq_laneq_f32(C11, B1, A0, 1);
  C20=vfmaq_laneq_f32(C20, B0, A0, 2); C21=vfmaq_laneq_f32(C21, B1, A0, 2);
  C30=vfmaq_laneq_f32(C30, B0, A0, 3); C31=vfmaq_laneq_f32(C31, B1, A0, 3);
}
// Store final Cr omitted for brevity
\end{lstlisting}
\end{minipage}
\end{center}
\vspace*{-2ex}
\caption{Packing (top) and \axpy-oriented micro-kernel (bottom) for $\mr \times \nr = 4 \times 8$, 
FP32 data, and  VL=128 bits in ARMv8.2 NEON. 
In the top figure, $\odot$ refers to an \axpy
between the left-hand side scalar operand and the right-hand side vector register.%
}
\label{fig:gotoblas2_micro}
\end{figure*}

Figure~\ref{fig:gotoblas2_micro} illustrates the GotoBLAS2 packing scheme of $A_c$ and $B_c$ for \axpy-oriented architectures along with a simple example micro-kernel, for ARMv8.2 NEON, operating on an $\mr \times \nr = 4 \times 8$ micro-tile $C_r$. 
In this particular case, $\mc=24$, $\nc=16$ and
the assumed baseline data type is FP32, with ARMv8.2 NEON VL=128-bit vector registers capable of holding four FP32 values each.

The above ARMv8.2 NEON code leverages the data type
\texttt{float32x4\_t} for variables representing vectors of four FP32 elements,
plus vector intrinsics for
vector load (\texttt{vld1q\_f32}), vector store (\texttt{vld1q\_f32}), and vector \axpy operations 
(\texttt{vfmaq\_ laneq\_f32}). 
With respect to the last one,
given three input vectors \texttt{v0}, \texttt{v1}, \texttt{v2}, each comprising 4 FP32 numbers,
\begin{lstlisting}[style=Ccode,morekeywords={float32x4_t,const}]
 float32x4_t vfmaq_laneq_f32(float32x4_t v0, 
                             float32x4_t v1, 
                             float32x4_t v2, 
                             const int lane)
\end{lstlisting}
computes the \axpy
\begin{lstlisting}[style=Ccode,caption={}]
 vr[i] = v0[i] + v1[i] * v2[lane], i=0,1,2,3.
\end{lstlisting}
~\\
and updates the output vector \texttt{vr}, which also contains 4 FP32 numbers.
Lines 5--8 in the code load the contents of $C_r$ into 8 vector registers leveraging 
the macro \texttt{Cr} (see line~1) to access the elements of the row-wise stored matrix~$C$.
Inside the loop,
lines 11--12 retrieve the $k$-th column and row of $A_r,B_r$ into 1+2 additional
vector registers; and lines 13--16 decompose the iteration rank-1 update into eight \axpy operations,
matching the type of instructions supported by ARMv8.2 NEON's ISA.

More advanced micro-kernels adhere to the same basic principles yet generally select larger micro-tile sizes ($\mr \times \nr$), apply loop unrolling, and/or leverage software pipelining to further improve performance~\cite{Dowd98}.
Furthermore, the micro-kernel is implemented in C with vector intrinsics that the compiler directly maps
into hardware-supported vector instructions or, sometimes, in assembly to accommodate low-level software pre\-fetching instructions. 

To close this presentation of the original micro-kernel,
Table~\ref{tab:intrinsics} maps the ARMv8.2 NEON vector intrinsics in Figure~\ref{fig:gotoblas2_micro} 
to ARM SVE (Scalable Vector Extension) and RISC-V RVV1.0.\\

\begin{table}
{\scriptsize
\begin{center}
\begin{tabular}{llll}
Operation & ARMv8.2 NEON & ARM SVE & RVV 1.0 \\
\toprule
Load & \texttt{vld1q\_f32} & \texttt{svld1\_f32} & \texttt{vle32\_v\_f32m1} \\
Computation &\texttt{vfmaq\_laneq\_f32} & \texttt{svmla\_lane\_f32} & \texttt{vfmacc\_vf\_f32m1} \\
Store & \texttt{vst1q\_f32} & \texttt{svst1\_f32} & \texttt{vse32\_v\_f32m1} \\
\bottomrule
\end{tabular}
\end{center}
\caption{Vector intrinsics equivalence between ARMv8.2 NEON, ARM SVE and RISC-V RVV 1.0 for FP32.}\label{tab:intrinsics}
}
\end{table}

In the following two sections, we use several examples to 
describe how the ISA and hardware have heterogeneously evol\-ved to meet the demands of DL inference workloads for MIP arithmetic.
Across all these cases, the overall structure of the \gemm operation remains consistent. Specifically, 
as discussed in Section~\ref{sec:gemm} the
macro-kernel is organized around five tiling loops, two packing routines that copy selected blocks of the input matrices into micro-panels, and a micro-kernel. 
Moreover, the micro-kernel itself follows a common algorithmic template: it preloads the output micro-tile 
$C_r$ into processor vector registers, iteratively updates 
$C_r$ within the loop body using the contents of the micro-panels 
$A_r,B_r$, and writes the updated result back to memory once computation is complete.

The key differences lie in the hardware-specific organization of the data into the packed buffers 
$A_c,B_c$,
and in the implementation of the micro-kernel using low-level MIP instructions specific 
to each ISA/architecture.
\section{Micro-kernel for \dotp-oriented SIMD Units and MIP}
\label{sec:mip-simd}

Integer arithmetic is highly relevant for quantized DL inference, particularly in production and edge environments, because it enables faster and more efficient execution of models with lower power and memory requirements. 
In addition, casting \gemm in terms of \dotp instead of \axpy is better suited to quantized 
DL inference because it preserves the linear structure, 
simplifies scale fusion, avoids overflow and scaling mismatch due to bias addition, and 
maintains precision and correctness across fused operations~\cite{natesh2025pqsprunequantizesort,finkelstein2019fightingquantizationbiasbias}.
In practice, the \dotp kernels are often computed using MIP arithmetic with 
the multiplication of inputs 
(weights and activations) in INT8 and the result accumulated in INT32.
This provides a larger dynamic range, preserving the integrity of the results even when using low-precision inputs,
and maintains compute and memory bandwidth in check~\cite{10.5555/3618408.3619993,10.5555/3600270.3602468}.

In this section we discuss in detail the ISA support for \dotp-based MIP arithmetic 
and the effect on the configuration of the micro-kernel and packing scheme 
using two mature examples:
ARMv8.0-A NEON 
and ARMv8.2 NEON, 
both with VL=128 bits.
In addition, we extend the design of the \dotp-oriented solution to ARM SVE2. 

\subsection{ARMv8.0-A NEON}
In order to formulate a MIP realization for this SIMD ISA, 
we address the rank-1 update inside loop \texttt{L6} of the micro-kernel with a combination of the following three vector intrinsics:
\begin{lstlisting}[style=Ccode,morekeywords={int16x8_t,int8x8_t,int32x4_t,int8x8_t}],caption={}]
 int16x8_t vmull_s8(int8x8_t     v0, int8x8_t v1)
 int16x8_t vmlal_s8(int16x8_t    v0, int8x8_t v1, 
                                     int8x8_t v2)
 int32x4_t vpadalq_s16(int32x4_t v0,int16x8_t v1)
\end{lstlisting}
The first intrinsic takes two input vectors 
\texttt{v0}, \texttt{v1}
with eight INT8 numbers each, and performs their element-wise
multiplication, returning the result on a destination vector 
\texttt{vr} with
eight INT16 numbers:
\begin{lstlisting}[style=Ccode,caption={}]
 vr[i] = v0[i] * v1[i], i=0,1,...,7.
\end{lstlisting}
~\\
The second intrinsic takes 
an input vector
\texttt{v0} with eight
INT16 numbers, and
two input vectors 
\texttt{v1}, \texttt{v2} 
with eight INT8 numbers each, 
and computes a result \texttt{vr} of the same type as \texttt{v0}, with:
\begin{lstlisting}[style=Ccode,caption={}]
 vr[i] = v0[i] + v1[i] * v2[i], i=0,1,...,7.
\end{lstlisting}
~\\
The last intrinsic receives two input vectors
\texttt{v0}, \texttt{v1}, with 
four INT32 numbers the first one and 
eight INT16 numbers the second, and returns
an output 
\texttt{vr}
consisting of four INT32 numbers as follows:
\begin{lstlisting}[style=Ccode,caption={}]
 vr[i] = v0[i] + v1[2*i] + v1[2*i+1], i=0,1,2,3.
\end{lstlisting}
~\\

\begin{figure*}[t!]
\begin{center}
\includegraphics[width=0.9\textwidth]{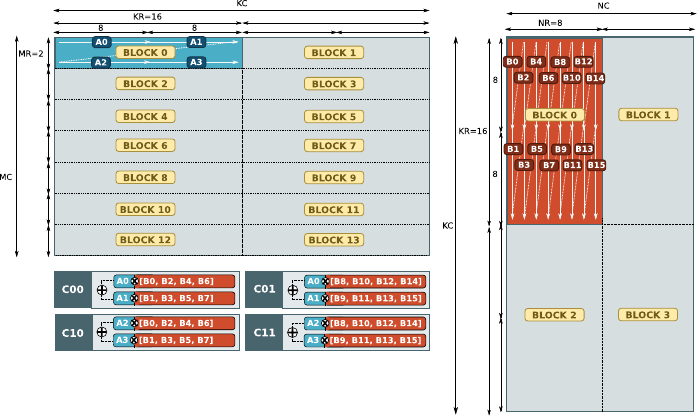}
\centering
\begin{minipage}{0.95\linewidth}
\begin{lstlisting}[linewidth=\textwidth, language=C,alsoletter={.},deletekeywords={.sum},morekeywords={v4_fp32,float32x4_t},breaklines=true]
int8x8_t   A0,  A1, A2,  A3, B0,  B2, B1,  B3;
int16x8_t  T00, T01, T10, T11;
int32x4_t  C00, C01, C02, C03, C04, C05, C06, C07, C10, C11, C12, C13, C14, C15, C16, C17;

for (k=0; k<kc; k+=kr) { // Loop L6, with stride kr=16
  A0=vld1_s8(&Ar[baseA+0]); A2=vld1_s8(&Ar[baseA+16]); 
  A1=vld1_s8(&Ar[baseA+8]); A3=vld1_s8(&Ar[baseA+24]); baseA+=32; 
  B0=vld1_s8(&Br[baseB+0]); B2=vld1_s8(&Br[baseB+16]); 
  B1=vld1_s8(&Br[baseB+8]); B3=vld1_s8(&Br[baseB+24]); baseB+=32;
  // Compute the product of the micro-panel [A0, A1; A2, A3]
  // with columns [[B0; B1], [B2; B3]]
  T00=vmull_s8(A0, B0);      T01=vmull_s8(A0, B2);      T10=vmull_s8(A2, B0);      T11=vmull_s8(A2, B2);
  T00=vmlal_s8(T00, A1, B1); T01=vmlal_s8(T01, A1, B3); T10=vmlal_s8(T10, A3, B1); T11=vmlal_s8(T11, A3, B3);
  C00=vpadalq_s16(C00, T00); C01=vpadalq_s16(C01, T01); C10=vpadalq_s16(C10, T10); C11=vpadalq_s16(C11, T11);
  // Remaining part of loop body omitted for simplicity
}
// Omitted for simplicity: Reduce the results on [C00, C01,...,C07, C10, C11,...,C17]
// accumulating the result on micro-tile [C00, C01; C10, C11]
\end{lstlisting}
\end{minipage}
\vspace*{-2ex}
\caption{Packing (top) and \dotp-oriented micro-kernel (bottom) for $\mr \times \nr = 2 \times 8$, 
MIP INT8+INT32 data, and  VL=128 bits in ARMv8.0-A NEON.
In the top figure, $\otimes$ refers to \dotp operations
between the left-hand side vector and each vector of the right-hand side;
$\oplus$ refers to the element-wise addition.
}
\label{fig:armv8.0_packing}
\end{center}
\end{figure*}

An excerpt of a sample micro-kernel with $\mr \times \nr = 2 \times 8$ as well as the corresponding packing scheme
for ARMv8.0-A NEON are
displayed in Figure~\ref{fig:armv8.0_packing}. 
At this point we remind that the purpose of stacking the data 
in the buffers $A_c,B_c$, following a certain pattern,
is to enable accessing their contents with stride-1 from the micro-kernel. 
In this line, 
the white arrows in the packing scheme specify the layout of the elements of $A_c,B_c$ in memory.
Inside the loop, lines 6--7 load the $2 \times 16$ micro-panel $A_r$ into four vector registers
(\texttt{A0}--\texttt{A3}); 
lines 8--9 load the first two columns of the $16 \times 8$ micro-panel $B_r$ using four additional vector
registers
(\texttt{B0}--\texttt{B3}); 
and lines 12--14 compute the product of these elements, leaving the result into a $2 \times 4$
micro-tile stored in registers
\texttt{C00}, \texttt{C01}, 
\texttt{C10}, \texttt{C11}. 
The loop repeats the same process in lines 8--14 three more times (omitted for brevity), 
in order to load the remaining six columns of $B_r$, 
two at a time, computing the product of the same micro-panel $A_r$ with the corresponding pair
yet leaving the result on a separate $2 \times 4$ micro-tile per column pair.
(e.g., \texttt{C02}, \texttt{C03}, 
 \texttt{C12}, \texttt{C13} for the second pair). 

Upon completion of the loop body, we obtain four separate micro-tiles, with $2 \times 4$ elements each, that
need to be reduced and concatenated into the final $2 \times 8$ micro-tile.
This is achieved in two passes via the
vector intrinsic:
\begin{lstlisting}[style=Ccode,morekeywords={int16x8_t,int8x8_t,int32x4_t,int8x8_t},caption={}]
 int32x4_t vpaddq_s32(int32x4_t v0, int32x4_t v1)
\end{lstlisting}
~\\
which receives two
input vectors 
\texttt{v0}, \texttt{v1}, with 
four INT32 numbers each, and computes an output
\texttt{vr} of the same type as follows:
\begin{lstlisting}[style=Ccode,caption={}]
 vr[0] = v0[0] + v0[1], vr[1] = v0[2] + v0[3],
 vr[2] = v1[0] + v1[1], vr[3] = v1[2] + v1[3].
\end{lstlisting}
~\\

To better understand the full process, let us
consider for example 
\texttt{C00}, which should contain $C_r(0,0\colon3)$ upon execution of the micro-kernel.
As shown in the box representing the micro-tile $C_r$ in the top part of Figure~\ref{fig:armv8.0_packing},
these entries are obtained from the \dotp 
\texttt{A0}$\otimes$\texttt{[B0, B2, B4, B6]} (partially computed with the element-wise multiplications via \texttt{vmull\_s8}) plus 
the \dotp 
\texttt{A1}$\otimes$\texttt{[B1, B3, B5, B7]} (partially computed with the element-wise multiplications and additions via \texttt{vmlal\_s8}).
The result is a vector with eight numbers that need to be reduced into four numbers
(computed via \texttt{vpadalq\_s16}).
Finally, as we obtain a $2 \times 4$ partial result per column pair of $B_r$, two additional reduction
rounds yield the sought-after result (attained with successive invocations to \texttt{vpaddq\_s32}).
In summary, the ARMv8.0-A NEON ISA allows to decompose the rank-1 update inside the loop \texttt{L6} into a collection
of \dotp products that match the ARMv8.0-A ISA but, as we exposed, the process is quite elaborate.

\subsection{ARMv8.2 NEON}
Among other features, ARMv8.2 NEON
offers a more direct support for decomposing the
rank-1 update inside loop~\texttt{L6} into a collection of \dotp operations.
Concretely, ARMv8.2 NEON introduces the vector intrinsic:
\begin{lstlisting}[style=Ccode,morekeywords={int32x4_t,int8x16_t,const}]
 int32x4_t vdotq_laneq_s32(int32x4_t v0, 
                           int8x16_t v1, 
                           int8x16_t v2, 
                           const int lane)
\end{lstlisting}
that, given two input vectors 
\texttt{v1,v2} each comprising 16 INT8 numbers,
plus a third input vector \texttt{v0} with 4 INT32 numbers, computes
\begin{lstlisting}[style=Ccode,caption={}]
 vr[i] = v0[i] + v1[i*4+0] * v2[lane*4+0]  
               + v1[i*4+1] * v2[lane*4+1]  
               + v1[i*4+2] * v2[lane*4+2]  
               + v1[i*4+3] * v2[lane*4+3], 
               i=0,1,2,3.
\end{lstlisting}
~\\

Figure~\ref{fig:armv8.0_packing} shows that the use of this MIP intrinsics allows a notable 
simplification of the micro-kernel with respect to that presented for ARMv8.0-A NEON. 
As in the previous cases, the figure also includes the corresponding packing scheme for the buffers
$A_c,B_c$ (with the data layout specified by means of white arrows)
that accommodates a stride-1 access to the elements of the micro-panels for the 
$\mr \times \nr = 4 \times 16$ sample micro-kernel in the figure.
Lines 5--8 load the micro-tile $C_r$ prior to the main loop and,
inside the latter, lines 10--11 load the micro-panel $A_r$.
For brevity, the code includes an inner loop (lines 12--20) that processes four 
lanes but this is unrolled in the actual implementation. 
In the innermost loop, lines 13--14 load a $4 \times 16$ block of the micro-panel $B_r$;
and lines 15--19 perform the arithmetic using the 
afore-mentioned 
vector intrinsic \texttt{vdotq\_laneq\_s32}.

\begin{figure*}[t!]
\begin{center}
\includegraphics[width=0.7\textwidth]{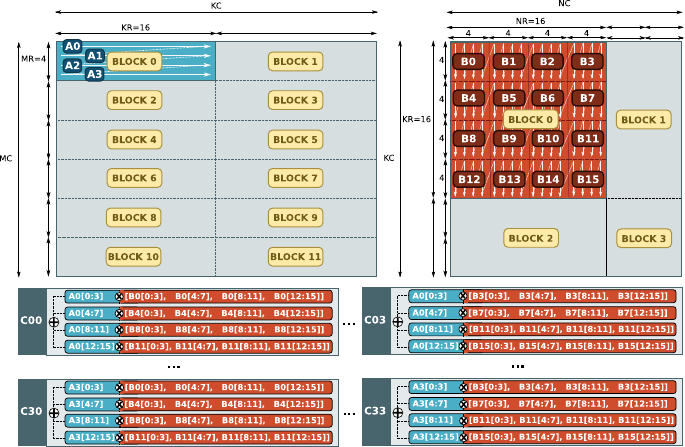}
\centering
\begin{minipage}{\linewidth}
\begin{lstlisting}[linewidth=\textwidth, language=C,alsoletter={.},deletekeywords={.sum},morekeywords={v4_fp32,float32x4_t},breaklines=true]
int32x4_t C00, C01, C02, C03, C10, C11, C12, C13, C20, C21, C22, C23, C30, C31, C32, C33;
int8x16_t A0, A1, A2, A3, B0, B1, B2, B3;

// Load initial Cr
vld1q_s32(&Cr(0,0), C00); vld1q_s32(&Cr(0,4), C01); vld1q_s32(&Cr(0,8), C02); vld1q_s32(&Cr(0,12), C03);
vld1q_s32(&Cr(1,0), C10); vld1q_s32(&Cr(1,4), C01); vld1q_s32(&Cr(1,8), C12); vld1q_s32(&Cr(1,12), C13);
vld1q_s32(&Cr(2,0), C20); vld1q_s32(&Cr(2,4), C01); vld1q_s32(&Cr(2,8), C22); vld1q_s32(&Cr(2,12), C23);
vld1q_s32(&Cr(3,0), C30); vld1q_s32(&Cr(3,4), C31); vld1q_s32(&Cr(3,8), C32); vld1q_s32(&Cr(3,12), C33);
for (int k=0; k<kc; k+=kr) { // Loop L6, with stride kr=16
  A0=vld1q_s8(&Ar[baseA+0]);  A1=vld1q_s8(&Ar[baseA+16]); A2=vld1q_s8(&Ar[baseA+32]); A3=vld1q_s8(&Ar[baseA+48]); 
  baseA+=64;
  for (int lane=0; lane<4; lane++) {
    B0=vld1q_s8(&Br[baseB+0]);  B1=vld1q_s8(&Br[baseB+16]); B2=vld1q_s8(&Br[baseB+32]); B3=vld1q_s8(&Br[baseB+48]); 
    baseB+=64;
    C00=vdotq_laneq_s32(C00, B0, A0, lane); C01=vdotq_laneq_s32(C01, B1, A0, lane); 
    C02=vdotq_laneq_s32(C02, B2, A0, lane); C03=vdotq_laneq_s32(C03, B3, A0, lane);
    C10=vdotq_laneq_s32(C10, B0, A1, lane); C11=vdotq_laneq_s32(C11, B1, A1, lane); 
    // Remaining dot products omitted for brevity
    C32=vdotq_laneq_s32(C32, B2, A3, lane); C33=vdotq_laneq_s32(C33, B3, A3, lane);
  }
// Store final Cr, omitted for simplicity
}
\end{lstlisting}
\end{minipage}
\vspace*{-2ex}
\caption{Packing (top) and \dotp-oriented micro-kernel (bottom) for $\mr \times \nr = 4 \times 16$, 
MIP INT8+INT32 data, and  VL=128 bits in ARMv8.2 NEON.}
\label{fig:armv8.2_packing}
\end{center}
\end{figure*}

Let us consider again the computation of 
\texttt{C00}, which should contain $C_r(0,0\colon3)$ upon completion of the micro-kernel.
The box for the micro-tile $C_r$ in the top part of Figure~\ref{fig:armv8.2_packing}
indicates that these four elements are obtained as the result of 16 \dotp operations, with four
computed per iteration of the innermost loop. For example,
for \texttt{lane}=1, 
the invocation of \texttt{vdot\_int8\_32} computes the four \dotp operations
\texttt{A0[4:7]}$\otimes$\texttt{[B4[0:3], B4[4:7], B4[8:11], B4[12:15]]}, adding the result
to the current contents of \texttt{C00}.

\subsection{ARM SVE2}

ARM SVE2 is a vector-length agnostic ISA that supports vector registers of flexible length, between 128 and 2048 bits.
This allows to write the code once, but run it optimally on CPUs with different vector lengths
without having to recompile it.

ARM SVE2 includes the following vector intrinsic for the \dotp:
\begin{lstlisting}[style=Ccode,morekeywords={svint32_t,svint8_t,svint8x4_t,const}]
 svint32_t svdot_lane_s32(svint32_t  v0, 
                          svint8_t   v1, 
                          svint8x4_t v2, 
                          const int  lane);
\end{lstlisting}
that, given an input vector 
\texttt{v1} with $4n$ INT8 numbers;
an input tuple 
\texttt{v2} of four vectors, each with
$4n$ INT8 numbers;
plus a third input vector \texttt{v0} with $n$ INT32 numbers, computes
\begin{lstlisting}[style=Ccode,caption={}]
 vr[i] = v0[i] + v1[i*4+0] * v2[lane][i*4+0]  
               + v1[i*4+1] * v2[lane][i*4+1]  
               + v1[i*4+2] * v2[lane][i*4+2]  
               + v1[i*4+3] * v2[lane][i*4+3], 
               i=0,1,...,n-1.
\end{lstlisting}
~\\

For a CPU with VL=128 bits, the packing scheme in ARM SVE2 for an 
$\mr \times \nr = 4 \times 8$ micro-kernel with MIP INT8+INT32 data is basically the same 
given in Figure~\ref{fig:armv8.2_packing}.
A vector-length agnostic micro-kernel is rather more complex though and its presentation
is omitted for brevity.\\

In the next section we show how matrix engines further simplify the micro-kernel code 
yet require elaborate packing schemes.
\section{Micro-kernel for Tile-oriented Matrix Engines and MIP}
\label{sec:mip-mengine}

Matrix multiplication units, or engines, have rapidly evolved into highly specialized accelerators tailored for DL and scientific computing. NVIDIA was among the first to mainstream this shift with the introduction of Tensor Cores in its Volta architecture, 
enabling massively parallel mixed-precision matrix operations. Building on this trend, Intel's AMX 
brought similar capabilities to x86 CPUs, introducing a tile-based compute model with new ISA-level support 
especially optimized for 
data formats common in DL inference. 
Additionally, ARM has incorporated matrix acceleration directly into its architecture through 
SME 
for the Cortex-A cores. 
Across all these platforms--NVIDIA, Intel, and ARM--the focus is on tightly integrating matrix engines into the execution pipeline, scaling performance via parallelism, dedicated hardware pathways, and ISA-level enhancements. This evolution reflects the growing importance of matrix operations in modern workloads, and the need for architectural specialization to sustain performance and energy efficiency at scale.

In this section we illustrate how the \gemm micro-kernel and packing scheme are adapted for MIP to
three distinct matrix multiplication units: SpacemiT K1, Intel AMX and ARM SME.

\subsection{SpacemiT K1}

The Banana Pi BPI-F3 is a single-board computer with an 8-core, 64-bit SpacemiT K1 system-on-chip. The processor
supports 64GCVB RISC-V standard and RVV 1.0 with VL=256-bit vector registers. 
In addition, the processor integrates four 
IMEs 
specifically designed to accelerate MIP \gemm. The Spa\-cemiT K1 can off-load part of the arithmetic
to the IME via, e.g., the assembly instruction
\begin{lstlisting}[style=Ccode,morekeywords={}]
 vmadot v0, v10, v11 
\end{lstlisting}
which computes a small, MIP \gemm of dimensions $(m,n,k)=(4,4,8)$. %
Concretely, given 
a $4 \times 8$ input tile $A_r$ with INT8 elements stored in row-major order in the vector register \texttt{v10} and
an $8 \times 4$ input tile $B_r$ with INT8 elements stored in column-major order in \texttt{v11},
\texttt{vmadot} computes the product $C_r\,+\!\!= A_rB_r$, producing a $4 \times 4$ output tile $C_r$ with INT32 elements stored
row-wise in vector registers \texttt{v0}, \texttt{v1}. 
As discussed next, this instruction is the building block upon which we construct our MIP \gemm for 
quantized DL inference on the SpacemiT K1.

\begin{figure*}[t!]
\begin{center}
\includegraphics[width=0.7\textwidth]{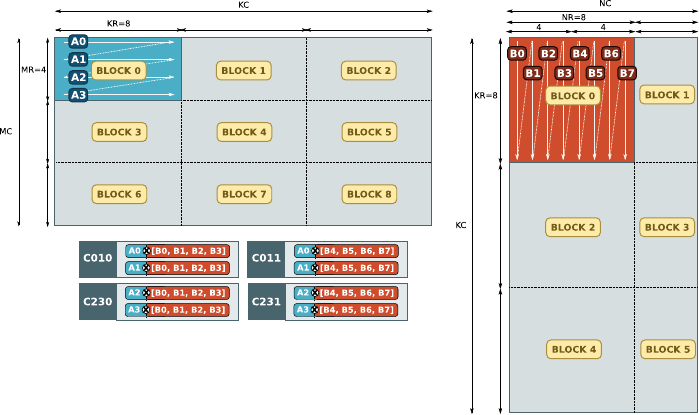}
\centering
\begin{minipage}{0.9\linewidth}
\begin{lstlisting}[linewidth=\textwidth, language=C,alsoletter={.},deletekeywords={.sum},morekeywords={v4_fp32,float32x4_t},breaklines=true]
int32x2x4_t C010, C230, C011, C231, tmp;
int8x4x8_t  A0123; 
int8x8x4_t  B0123, B4567; 

// Load initial Cr
vload_int32(&Cr(1,0), tmp); vslideup(tmp,C010); vload_int32(&Cr(0,0), C010);
vload_int32(&Cr(3,0), tmp); vslideup(tmp,C230); vload_int32(&Cr(2,0), C230);
vload_int32(&Cr(1,4), tmp); vslideup(tmp,C011); vload_int32(&Cr(0,4), C011);                     
vload_int32(&Cr(3,4), tmp); vslideup(tmp,C231); vload_int32(&Cr(2,4), C231);                     
// Micro-kernel
for (int k=0; k<kc; k+=kr) { // Loop L6, with stride kr=8
 vload_int8(&Ar[baseA], A0123); baseA+=32; 
 vload_int8(&Br[baseB], B0123); vload_int8(&Br[baseB+32], B4567); baseB+=64;
 C010,C230=vmadot_int8_32(A0123, B0123); C011,C231=vmadot_int8_32(A0123, B4567);
}
// Store final Cr, omitted for simplicity
\end{lstlisting}
\end{minipage}
\vspace*{-1ex}
\caption{Packing (top) and matrix-oriented micro-kernel (bottom) for $\mr \times \nr = 4 \times 8$, 
MIP INT8+INT32 data, and  VL=256 bits in the SpacemiT K1.}
\label{fig:ime_packing}
\end{center}
\end{figure*}

Given the access to the data from 
\texttt{vmadot}, the packing sche\-me for the SpacemiT K1 ISA
is shown in Figure~\ref{fig:ime_packing}, which also displays a high level
formulation of the micro-kernel using pseudo-instructions. The actual implementation of this component
is done in assembly as the accelerator does not support vector intrinsics.
Consider the $\mr \times \nr = 4 \times 8$ micro-kernel in the
figure. Line~5 loads  the first two rows of the left-hand side $4 \times 4$ block of the micro-tile $C_r$ 
(with INT32 elements) into the 
256-bit vector register \texttt{C010} (for \texttt{C00} and \texttt{C10}), with row~1 occupying the highest 128 bits and row~0 in
the lowest 128 bits.
Line~6 repeats the sequence to load rows~2 and~3 of the left block of the micro-tile into 
\texttt{C230} (for \texttt{C20} and \texttt{C30}); and lines~8--9 complete the process for the $4 \times 4$ right block of the micro-tile loading
its four rows into
\texttt{C011} and
\texttt{C231}.

The micro-kernel loop spans lines 10--16, processing $k_r=8$ rank-1 updates per iteration.
For this purpose, the micro-kernel loads 32 INT8 elements of $A_c$ into 
the vector register 
\texttt{A0123} (line~12); 
loads 2$\times$32 INT8 elements of $B_c$ into
the vector registers
\texttt{B0123} and
\texttt{B4567} (line~13); and
invokes twice the macro 
\texttt{vmadot\_int8\_32} which is a high level abstraction of the SpacemiT K1 instruction
\texttt{vmadot} (line 14).

The box for the micro-tile $C_r$ in the top part of Figure~\ref{fig:ime_packing} shows that upon completion 
of the micro-kernel execution,
for example,
the 256-bit vector register
\texttt{C010} mathematically contains the result of eight \dotp operations:
\texttt{A0}$\otimes$\texttt{[B0, B1, B2, B3]} (row~0 of the result) in the lowest 128 bits and
\texttt{A1}$\otimes$\texttt{[B0, B1, B2, B3]} (row~1 of the result) in the highest 128 bits.
We note that whether the matrix product between the contents of 
[\texttt{A0}; \texttt{A1}] and \texttt{[B0, B1, B2, B3]} are obtained via \dotp kernels or by some other sort
of building blocks is up to the hardware implementation of the SpacemiT K1 matrix multiplication engine.

\subsection{Intel AMX}

Intel AMX provides eight tile registers, with a capacity of 1024 bytes each, that can be used, for example,
to hold 256 INT32 values or 1024 INT8 numbers. This ISA 
can be utilized to implement the MIP with INT8+INT32 data via the 
intrinsic
\begin{lstlisting}[style=Ccode,morekeywords={constexpr}]
 void _tile_dpbssd (constexpr int c, 
                    constexpr int a, 
                    constexpr int b);
\end{lstlisting}
where 
\texttt{c} is the index of the tile register that will contain the output upon execution of the intrinsic,
and 
\texttt{a},  
\texttt{b} denote the indices of the tile registers for the input operands $A_r,B_r$. 
The above intrinsic computes a MIP INT8+INT32
\gemm with $(m,n,k)=(16,16,64)$. For this purpose, 
the tile registers \texttt{c} and \texttt{a} contain the corresponding tile operands in row-major order,
while the first column of \texttt{b} (64 elements) is stored in row-major order using the first four columns of the tile register
for $B_r$ (viewed as a $16 \times 64$ micro-panel); 
the second column of $B_r$ is stored in row-major order using the next four columns of the tile register; etc.

The packing scheme of an $\mr \times \nr = 16 \times 16$ MIP micro-kernel for Intel AMX is equivalent to that
of ARMv8.2 in Figure~\ref{fig:armv8.2_packing}, with $\mr=16$ rows instead of $4$, 
and $\kr=64$ instead of 16. The micro-kernel is rather simple, and basically
includes a load of the micro-tile $C_r$ prior to the
main loop (\texttt{L6}) that iterates over the $\kc$ dimension in steps of $\kr=64$. At each iteration,
the micro-kernel loads a micro-panel from $A_c$ and a micro-panel from $B_c$ and performs the tile
multiplication with a single invocation to the intrinsic \texttt{\_tile\_dpbssd}; upon completion,
the micro-tile $C_r$ is moved back to memory. The code is omitted for brevity.

\subsection{ARM SME}

Similarly to Intel's AMX, ARM SME has been recently introduced as
an architecture and ISA extension to provide support for matrix operations. SME is built on top of SVE and SVE2, adding
capabilities for matrix computation such as outer product instructions between two vectors, a matrix tile storage,
load/store instruction to/from tiles, and on-the-fly transposition, among others.

SME operates in the so-called {\em Streaming SVE} mode, in which 32 vector registers (Z0 to Z31) are provided.
The {\em Streaming SVE} mode determines a vector length (SVL) that does not necessarily match the vector length of
the Non-streaming (classic) SVE mode.
A new SME ZA ({\em Z Array}) storage, a bi-dimensional byte array of dimension SVL $\times$ SVL bytes, is also exposed.
As an example, for an SVL of 512 bits (as in Apple M4's AMX), the size of the ZA storage is $64 \times 64 = 4096$ bytes.

The ZA storage can be accessed as SVL-bit vectors that contain 8-, 16-, 32-, 64- or 128-bit
elements. The number of array vectors matches the number of bytes in SVL (e.g. if SVL, is 512-bit, the number of ZA array vectors is 64).
More interestingly, the ZA storage can be accessed as a set of tiles. A ZA tile is a square, bi-dimensional sub-array of elements of the
ZA storage. The width of a ZA tile is always SVL bits (e.g. 512 in our example), but the number of available tiles is fixed by the element
data type size; see Table~\ref{tab:sme_tiles}.

\begin{table}[h!]
\centering
\scriptsize
\begin{tabular}{@{}lll@{}}
\toprule
Element data type size & Number of tiles & Tile names         \\ \midrule
8-bit                  & 1               & {\sc ZA0.B}        \\
16-bit                 & 2               & {\sc ZA0.H-ZA1.H}  \\
32-bit                 & 4               & {\sc ZA0.S-ZA3.S}  \\
64-bit                 & 8               & {\sc ZA0.D-ZA7.D}  \\
128-bit                & 16              & {\sc ZA0.Q-ZA15.Q} \\ \bottomrule
\end{tabular}
\caption{Z array setups in ARM SME.}\label{tab:sme_tiles}
\end{table}

As an example, if the SVL is 512-bit (64-byte) and the element data size is 8-bit, the ZA storage can be viewed either as a single tile (named {\sc ZA0.B})
or as $64 \times (64 \times 1$-byte) vectors.
If the data type size is 16-bit, ZA can be viewed as two tiles ({\sc ZA0.H} and {\sc ZA1.H}), with each tile being
$16 \times (16 \times 2$-byte) vectors. A ZA tile can be accessed either as a whole, or as tile slices as groups of vectors
in a horizontal or vertical fashion.

Armed with these architectural state, in the integer flavor, SME provides outer product and accumulate instructions for
INT8 accumulating to INT32, and 16-bit integers accumulating to INT32 or INT64.
In the floating point counterparts, functionality is provided for FP16, BF16 and FP32 accumulating to either FP32 or FP64.
At a glance, these instructions calculate the outer product of two vectors in two SVE2 vector registers ({\sc Zn} and {\sc Zm}),
accumulate the result array with existing data in a ZA tile, and save the result to the same ZA tile.


%

For the case of MIP, taking as an example the INT8+INT32 variant, the instruction

\begin{lstlisting}[style=Ccode,morekeywords={}]
smopa <ZAda>.S, <Zn>.B, <Zm>.B
\end{lstlisting}
calculates the sum of four INT8 outer products, widens the results into INT32, and descructively adds the result to the 
destination tile. Note that, given that the input operands are considered as vectors (with no need of further reorganization), an \axpy-based micro-kernel would only need minor modifications to load/store the micro-tile of $C$ to/from the
ZA array, which greatly improves programmability and enables a quasi transparent port of existing micro-kernels.
\section{Experimental Evaluation}
\label{sec:experiments}

In this section, we assess the performance gains that can be attained by exploiting MIP arithmetic 
on specialized SIMD units as well as matrix engines for the type of workloads that are common in quantized DL inference.
For this purpose, we focus the evaluation on three platforms with distinct arithmetic units; and two representative DL cases, respectively arising in computer vision and natural language processing.

\subsection{Hardware and software setup}

\begin{table}
{\scriptsize
\renewcommand{\arraystretch}{1.2}
\setlength{\tabcolsep}{1.2pt}
\begin{center}
\begin{tabular}{lllrrr}
 Board                    & Processor        & ISA Architecture        & Freq. & \#Cores & RAM  \\ 
                          &                  & (Instruction Set)       & (GHz) &         & (GB) \\ \midrule
Raspberry Pi              & ARM Cortex-A72   & ARMv8.0-A NEON          & 1.80  & 4       & 8    \\
NVIDIA Jetson AGX Orin    & ARM Cortex-A78AE & ARMv8.2 NEON            & 2.20  & 12      & 64   \\
Banana Pi BPI-F3          & SpacemiT K1      & RVV 1.0                 & 1.60  & 8       & 4    \\
\bottomrule
\end{tabular}
\end{center}
\caption{Target systems evaluated in the experimental study.}\label{tab:boards}
}
\end{table}

Table~\ref{tab:boards} lists
the integrated CPUs and system boards used in the evaluation.
Given our focus on
inference with CPUs, in our experiments we do not utilize the
graphics processing unit (GPU) that is present in the NVIDIA Jetson AGX Orin.
While we recognize the interest of using GPUs for DL tasks, especially for 
model training, CPUs are still mainstream for DL inference due, among other factors, to 
power, size and cost constraints; 
thermal management issues; and strict consistent latency requirements.

We run our codes using 
all cores in each platform, with one thread per core.
At this point, we remark that the BPI-F3 contains~8 cores, but there are only
four IMEs attached to them. 
For this particular platform, we also report results using 4~cores and 4~IMEs.

In all the experiments,
the processor frequency is fixed; each test is run for at least 50
seconds; and the reported results correspond to average values
over these executions. 
The codes are compiled with the GNU
C compiler --\texttt{gcc} 10.5.0 on Raspberri Pi and NVIDIA Jetson AGX Orin; and 
and \texttt{gcc} 13.2.0 on Banana Pi BPI-F3-- with the optimization
flag \texttt{-O3} in both cases.

\subsection{Workload}

For evaluation, we focus on two representative types of DL models: 
\begin{itemize}
\item ResNet50v1.5 model + ImageNet dataset~\cite{he2016deep,8114708,Krizhevsky:2012:ICD:2999134.2999257}.
This is a residual network consisting of 50~layers and ~25.6 million parameters. It comprises
49 convolutional layers organized into 4 stages of residual blocks and 16~bottleneck blocks. The network also includes
a couple of pooling layers plus a final fully connected transform. {\color{black} As many of the convolutional layers share the same problem dimensions, for simplicity, when evaluating the layers separately we only report one size dimension for each group of identical layers, resulting in 20 different cases.} 
The arithmetic cost of the model is mostly due to the convolution operator:
\( 
O = \conv(F, I),
\label{eqn:conv}
\) 
where $I$ is a 4D input tensor consisting of $b$ input images of size  $\hi \times \wi \times \ci$ each,
$\hi \times \wi$ denote the image height $\times$ width, and
$\ci$  stands for its number of input channels (or features). Furthermore,
$F$ is a 4D filter tensor consisting of $\co$ kernels of
height $\times$ width $\times$ channels \textcolor{black}{$=\hf \times \wf \times \ci$} each. %
The operator 
then produces a 4D output tensor $O$, with $b$ outputs of size \textcolor{black}{ $\ho \times \wo \times \co$} each, 
where  $\ho \times \wo$ stand for the output height $\times$ width, and $\co$ for the number of output channels (which equals the number of filters).
To deal with this operator, we apply the lowering approach~\cite{Che06}, which transforms the tensor $I$ into an augmented
matrix via the function \textsc{im2col} and the convolution operator into a \gemm of dimensions
$(m,n,k) = (\co, b \ho \wo, \ci \hf \wf)$. 
The cost of this transform is included in the reported time.
\item BERT-Large model + SST-2 dataset~\cite{devlin-etal-2019-bert,socher-etal-2013-recursive}.
This is a transformer encoder
comprising an input embedding layer, 24 intermediate encoder blocks, and an output classification layer.
The tokens are represented using a vector of $d=1024$ embeddings; 
each encoder block 
consists of $h=16$ heads in the multi-head attention 
(\textsf{MHA}) module;
and the 
hidden dimension 
in the feed-forward network 
(\textsf{FFN}) module
is set to $f=4d$.
In our experiments, each of the $b$ input sentences contains $l=256$, 512 or 1024 tokens.
Given an encoder input $E_i$ of dimension $d \times lb$, and weight matrices
$W_Q,W_K,W_V,W_O,W_1,W_2$,
Table~\ref{tab:gemm}
displays the sequence of operations computed in an encoder block, identifying the dimensions of the
\gemm operations that concentrate the bulk of the computational cost of the full transformer encoder.
As all the encoder blocks are equal, we only report results for the execution of a single one.
Furthermore,
when evaluating the layers separately, we only report one result per \gemm of different size.
\end{itemize}

\begin{table}
\begin{center}
{\scriptsize
\renewcommand{\arraystretch}{1.2}
\setlength{\tabcolsep}{4pt}
\begin{tabular}{lrlccc}
       &         &                                  & $m$ & $n$ & $k$ \\ \hline
\textsf{MHA}   &  \textsf{M1-M3}. & $(Q,K,V)=(W_Q,W_K,W_V) \cdot E_I$ & $d$ & $lb$  & $d$ \\
       &  \textsf{M4}. & \textsf{Split}$(Q,K,V) \rightarrow 
       (Q^{i,j},K^{i,j},V^{i,j})_{i=1:h}^{j=1:b}$ & & & \\ 
       &     & for $j=1:b$ &     &       &       \\
       &     & ~~~for $i=1:h$ &     &       &       \\
       &  \textsf{M5}. & ~~~~~~$\bar{E}_1^{i,j}=((K^{i,j})^T \cdot Q^{i,j})/\sqrt{d_k}$ & $l$ & $l$   & $d/h$ \\
       &  \textsf{M6}. & ~~~~~~$E_1^{i,j}=$~\smax($\bar{E}_1^{i,j}$) &     &       &       \\
       &  \textsf{M7}. & ~~~~~~$E_2^{i,j}=V^{i,j} \cdot E_1^{i,j}$ & $d/h$ & $l$ & $l$ \\
       &  \textsf{M8}. & \textsf{Concatenate}$(E_2^{i,j})_{i=1:h}^{j=1:b} \rightarrow E_2$ &    &    &    \\
       &  \textsf{M9}. & $\bar{A}_O=W_O \cdot E_2$      & $d$ & $lb$ & $d$ \\
       & \textsf{M10}. & $A_O=$~\lnorm($\bar{A}_O+E_I$)        &    &      &    \\ \hline
\textsf{FFN}    & \textsf{F11}. & $\bar{E}_3=W_1 \cdot A_O$        & $f$ & $lb$   & $d$ \\
       & \textsf{F12}. & $E_3=$~\gelu($\bar{E}_3$)           &     &       &     \\
       & \textsf{F13}. & $\bar{E}_O=W_2 \cdot E_3$        & $d$ & $lb$   & $f$ \\
       & \textsf{F14}. & $E_O=$~\lnorm($\bar{E}_O+A_O$)        &    &      &     \\ 
\end{tabular}
}
\end{center}
\caption{Dimensions of the \gemm operations in the transformer encoder.}
\label{tab:gemm}
\end{table}

In both DL test cases we set the batch size $b=1$, corresponding to a latency-oriented single stream scenario.

\subsection{Quantization scheme}

Consider the standard \gemm operation $C = AB$, where $C$, $A$, and $B$ are floating-point matrices, and 
let $A^q, 
B^q$ 
respectively represent the quantized versions of $A,B$ with scaling factors $s_A, s_B$~\cite{10.5555/3122009.3242044}.
The previous \gemm operation can then be approximated as $C^r = s_As_B \,A^q B^q = s_As_B \,C^q \approx C$. Here, the primary 
computational cost lies in the integer-based \gemm operation $c^q=A^q B^q$, which leverages efficient integer arithmetic. In contrast, the scaling step with $(s_As_B)$ involves floating-point arithmetic 
yet contributes only a negligible cost~\cite{10.5555/3122009.3242044}.

For the residual network, we apply dynamic quantization to the convolution layers converted into \gemm via
the lowering approach. 
Following the general trend for transformers~\cite{10.5555/3600270.3602468},
for the BERT-Large model 
dynamic quantization is leveraged in
the weight-to-activation \gemm present in the encoder blocks 
(denoted as \textsf{M1}--\textsf{M3},
\textsf{M9},
\textsf{F11},
\textsf{F13} in Table~\ref{tab:gemm})
but not in their activation-to-activation 
counterparts 
(labeled as \textsf{M5},
\textsf{M7} in the same table).

{\color{black} Quantizing a neural network impacts the final model accuracy. The quantization effect over convolutional models has been studied 
in several works, such as~\cite{10.1007/978-3-030-41964-6_40}, demonstrating its low impact when using MIP INT8+INT32 arithmetic. 
In this work we also analyze the effect of INT8+INT32 quantization on the two selected DL models.

\subsection{Implementations of \gemm}

We evaluate two optimized implementations of \gemm:
\begin{itemize}
\item \textsf{FP32}: Our 32-bit floating point, \axpy-oriented version of \gemm that mimics 
    the ideas
    underlying the high performance implementation of matrix multiplication in GotoBLAS2
    (see Sections~\ref{sec:gemm} and~\ref{sec:simd}), with the cache configuration parameters
    $m_c,n_c,k_c$, the micro-kernel, and the parallel loop tailored 
    specifically for each platform. In a previous work~\cite{10.1007/978-3-031-69766-1_26} we demonstrated
    that, for the NVIDIA Jetson boards, our FP32 implementation of \gemm was competitive with or even outperformed those in 
    BLIS, OpenBLAS and ARM PL on ARM and RISC-V CPUs. 
\item \textsf{Q\_INT8\_INT32}: The MIP \gemm with that exploits the specialized hardware units in each platform,
    using the micro-kernels and packing schemes in Sections~\ref{sec:mip-simd} and~\ref{sec:mip-mengine}.
\end{itemize}    

When comparing the raw arithmetic throughput attained by these implementations, we report
the GOPS rate. 
In this sense, we consider that a \gemm of dimensions $(m,n,k)$ performs $2mnk$ ``useful'' operations, and a 
matrix-vector product involving
an $m \times n$ matrix requires $2mn$ useful operations.

{\color{black}
\subsection{ResNet50v1.5 model + ImageNet dataset}

\begin{figure*}[tbh!]
\centering
\includegraphics[width=0.98\columnwidth]{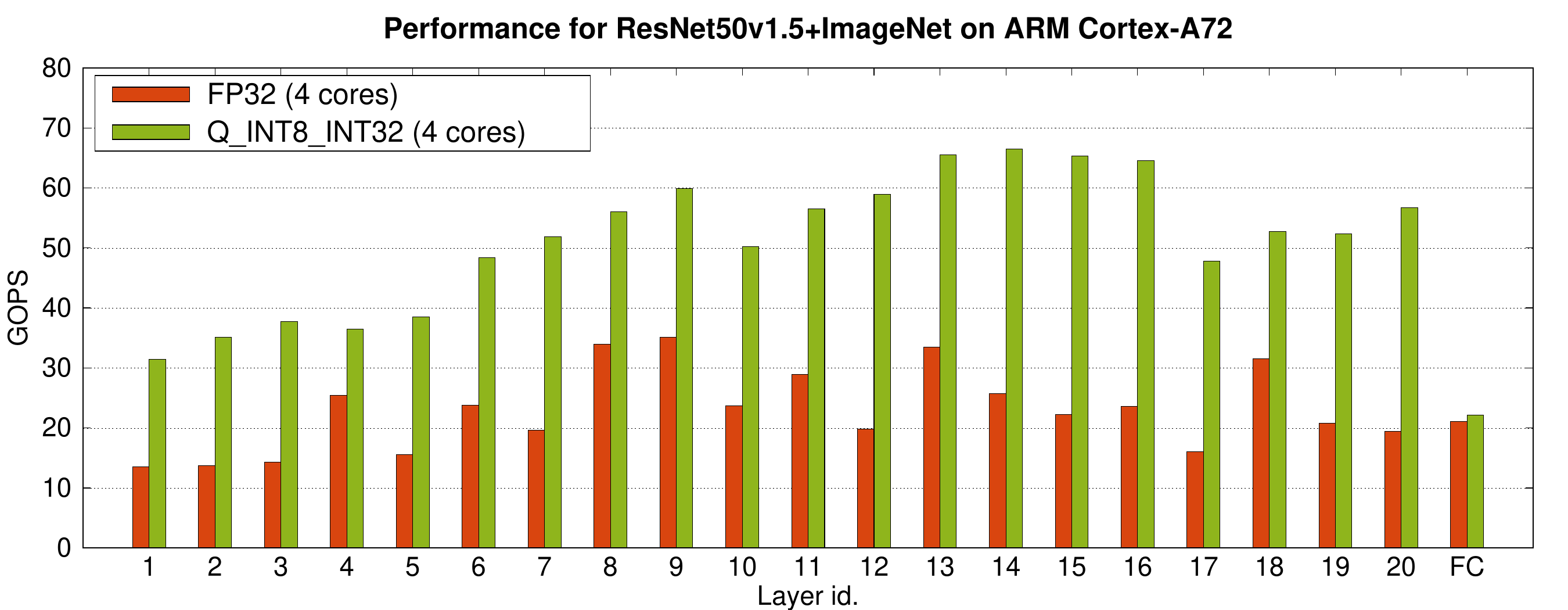} 
\includegraphics[width=0.98\columnwidth]{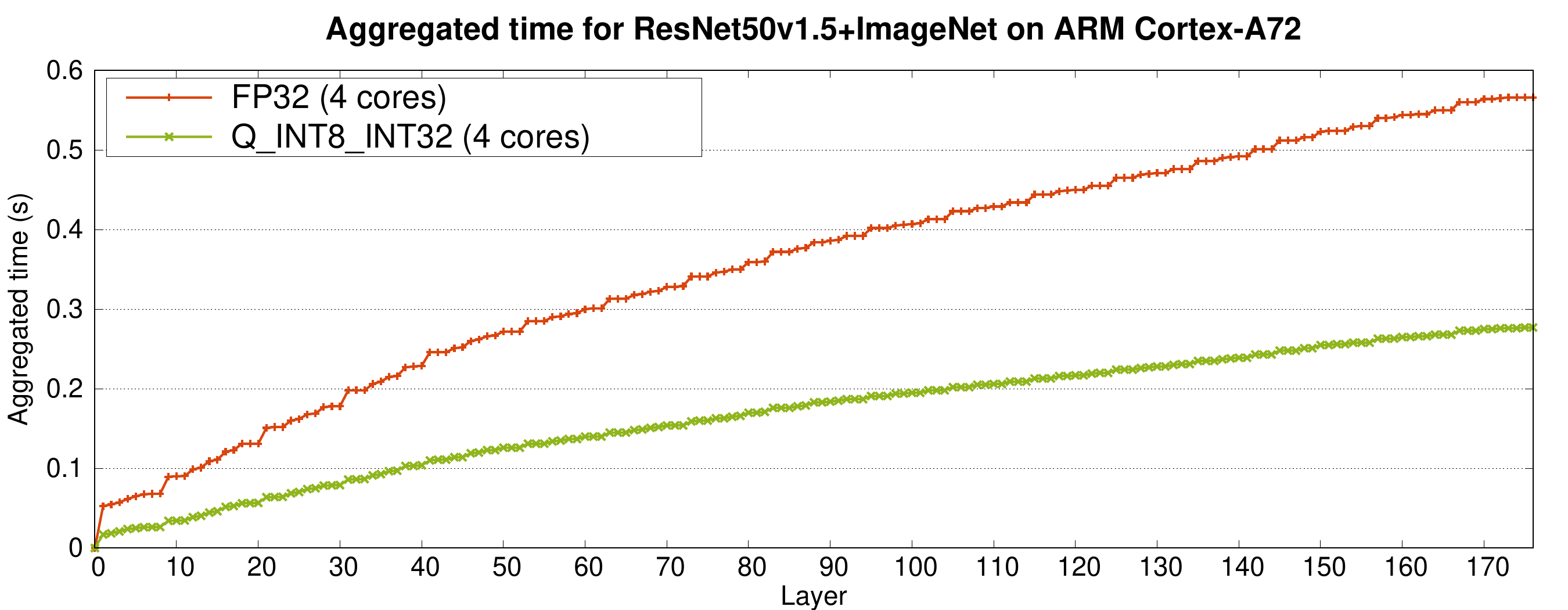} \\
\includegraphics[width=0.98\columnwidth]{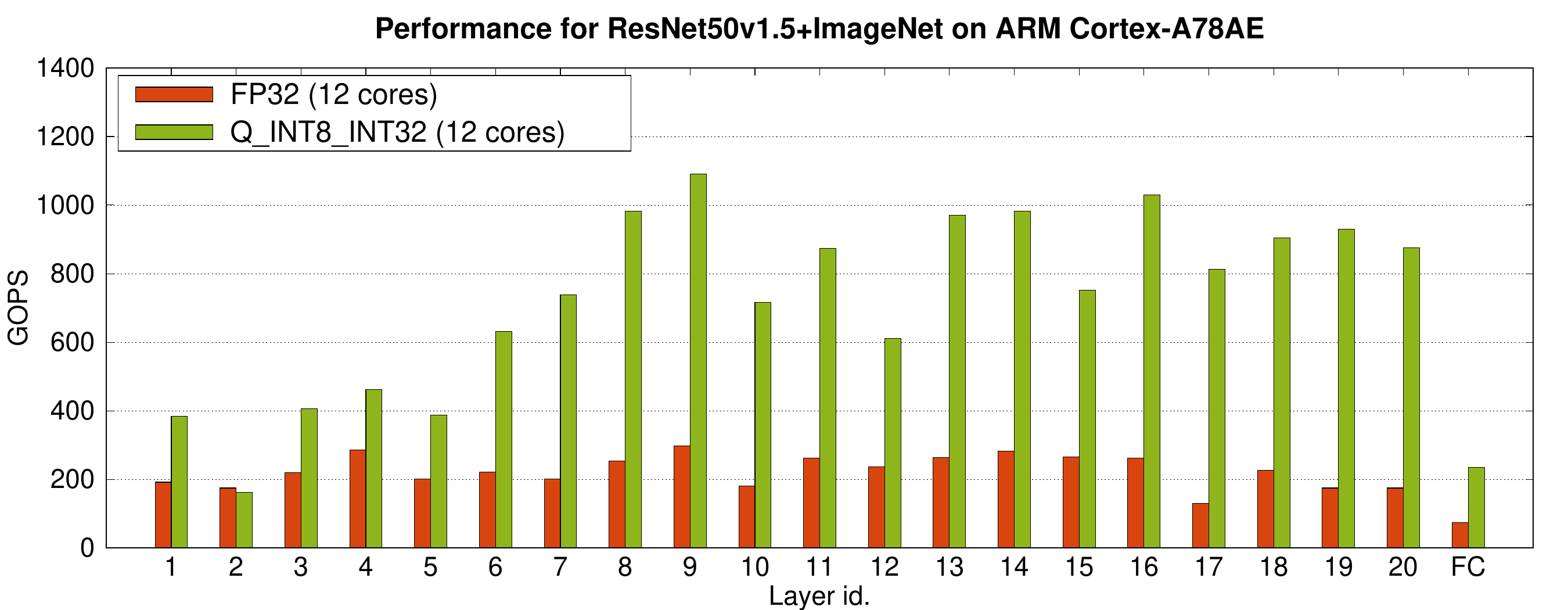} 
\includegraphics[width=0.98\columnwidth]{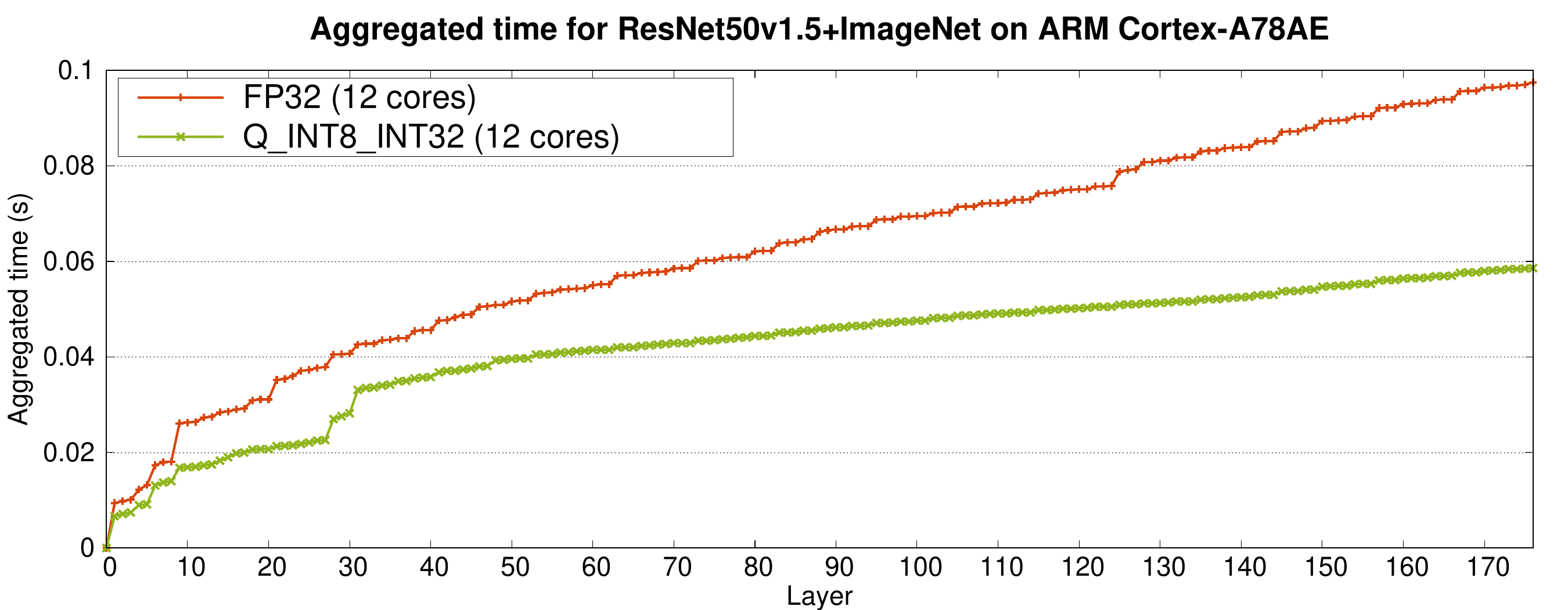} \\
\includegraphics[width=0.98\columnwidth]{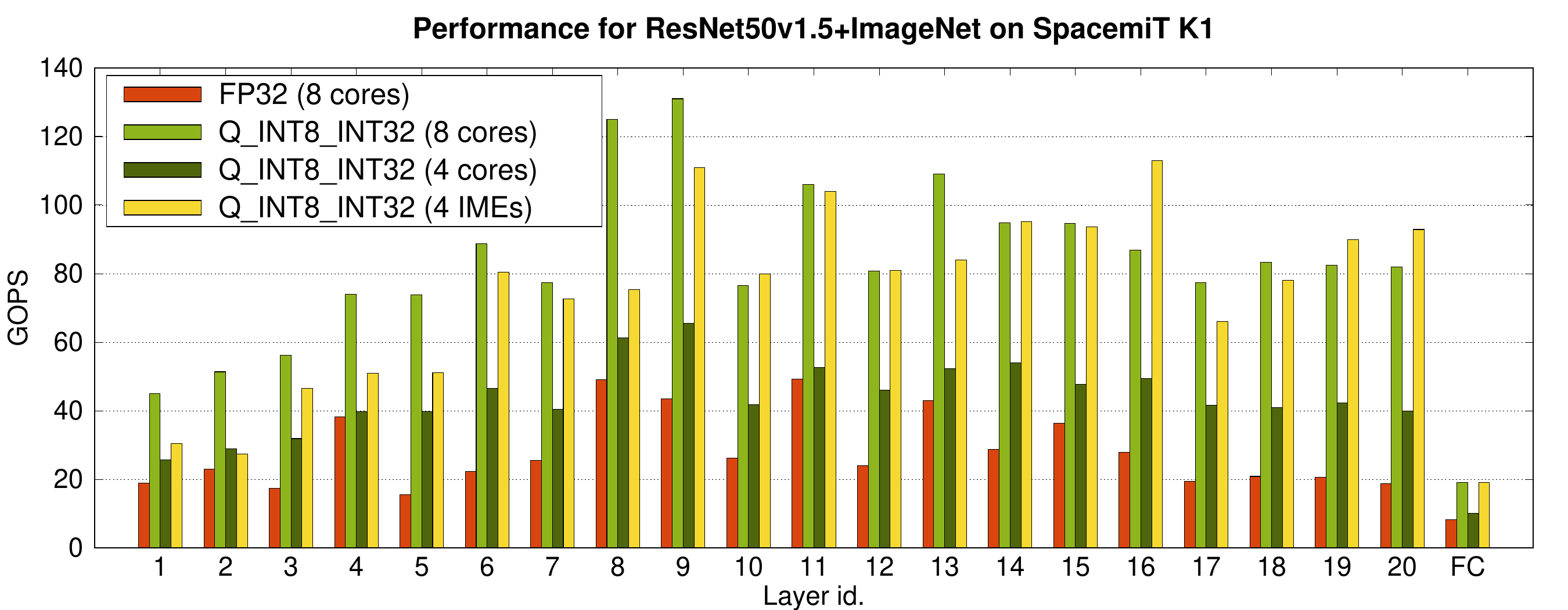} 
\includegraphics[width=0.98\columnwidth]{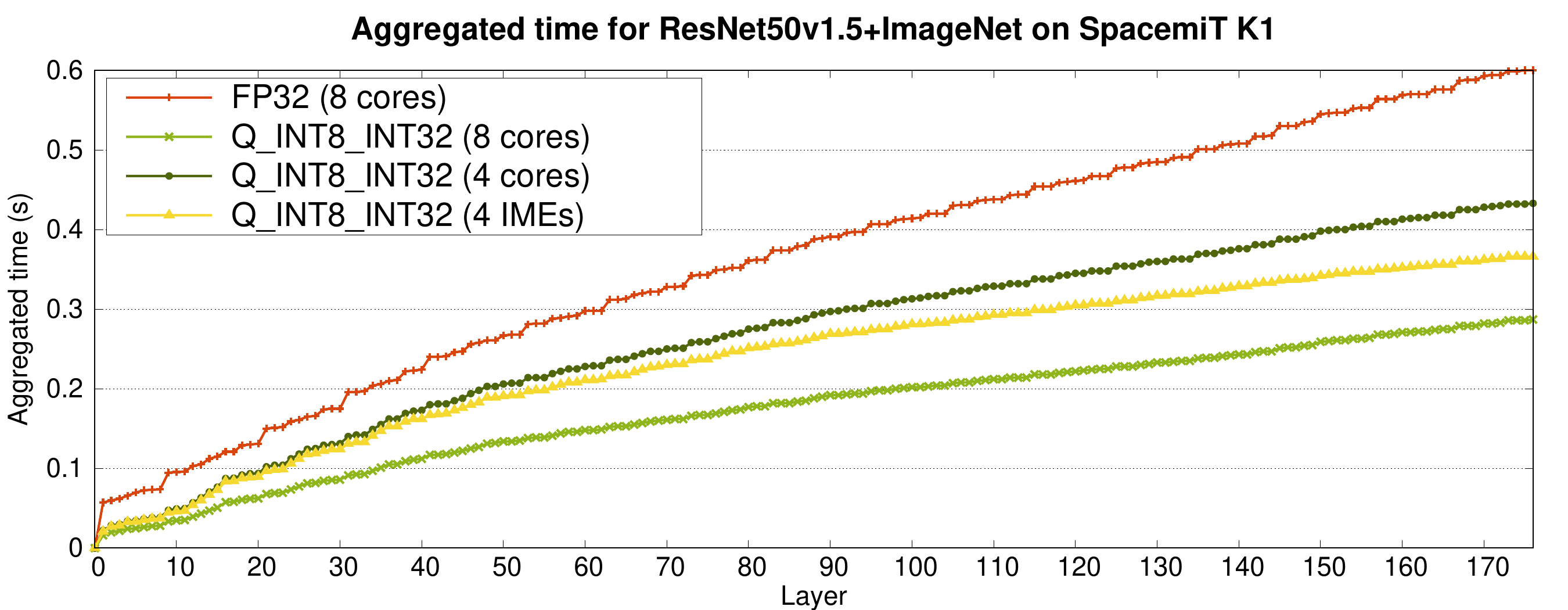} 
\caption{Performance of FP32 and MIP inference for ResNet50v1.5 model + ImageNet dataset. 
         Top row for ARM-Cortex A72, middle row for ARM-Cortex A78AE, and bottom row for SpacemiT K1. 
         Left column for individual layer performance (only layers of distinct sizes), and right column for aggregated time (all layers). }
\label{fig:resnet_performance}
\end{figure*}

Figure~\ref{fig:resnet_performance} evaluates the performance of the computer vision model
on the ARM Cortex-A72, ARM Cortex-A78AE, and SpacemiT K1 (top, middle, and bottom row, respectively). 
The left-hand side plots report the GOPS
measured for the \gemm resulting from the application of the lowering algorithm to the 20 distinct-size convolutions in this model (labeled as 1--20), 
plus that of the matrix-vector product in the final FC layer (labeled as \textsf{FC}). 
The right-hand side plots in the figure extends this study with the aggregated time of the entire model (all layers). 

Let us first consider the raw performance per layer (left-hand side column of plots in Figure~\ref{fig:resnet_performance}).
For the ARM-based platforms, the plots in the first two rows of plots demonstrate that the exploitation of MIP yields substantial benefits over the conventional FP32-based solution, with average speed-ups around 2.3$\times$ and 4.2$\times$ for ARM Cortex-A72 and ARM Cortex-A78AE, respectively, that matches the number of SIMD units in each processor. 
For the SpacemiT K1, the bottom row of plots includes results for executions using 4 CPUs and 4 IME devices as well. 
Taking the execution with all eight cores and FP32 arithmetic as reference,
the MIP execution with all the cores but no IMEs offers an average speed-up of 3.1$\times$. 
If we consider MIP arithmetic, the execution with the four IME accelerators outperform a run with 4 cores by a factor of 1.7$\times$. 
However, in case we use all eight cores and MIP arithmetic, the result is computed 1.2$\times$ faster than with the 4 IMEs.

For the aggregated time (right-hand side column in the same figure),
the speed-ups attained by the MIP solutions with respect to the reference FP32 configuration are 1.67$\times$, 2$\times$, and 2.1$\times$ for the 
ARM Cortex-A72, ARM Cortex-A78AE, and Space\-miT K1 with 8 cores, respectively.
\textcolor{black}{Regarding the speed-up for K1 with 4 cores and 4 IMEs, are 1.4$\times$ and 1.64$\times$ respectively.}

{\color{black}
Table~\ref{tab:resnet50_accuracy} shows the limited impact of the quantization process on the evaluation of the ResNet50v1.5 model with two subsets with different numbers of samples. The maximum accuracy drop is around 1\% when using the subset formed by 3600 random samples and similar results are observed for the 
larger test case.
}
\begin{table}[th!]
{\color{black} \small
\begin{center}

\begin{tabular}{lccc}
 \textbf{Model} & \textbf{\# Samples}  &\textbf{Top-1} & \textbf{Top-5} \\ 
               &                      &\textbf{Accuracy} & \textbf{Accuracy} \\ \midrule
 FP32& \multirow{2}{*} {~~3600}  &  0.752 & 0.920 \\
 Quantized& &  0.741 & 0.915 \\ 
\midrule
 FP32 & \multirow{2}{*} {50000} &  0.711 & 0.899 \\
 Quantized & &  0.705 & 0.895 \\ 
\bottomrule
\end{tabular}
\end{center}
\caption{\textcolor{black}{Accuracy of the original (FP32) and quantized (INT8+INT32) ResNet50v1.5 model on ImageNet for two validation set lengths.}} \label{tab:resnet50_accuracy}
}
\end{table}

\subsection{BERT-Large model + SST-2 dataset}

\begin{figure*}[t!]
\centering
\includegraphics[width=0.98\columnwidth]{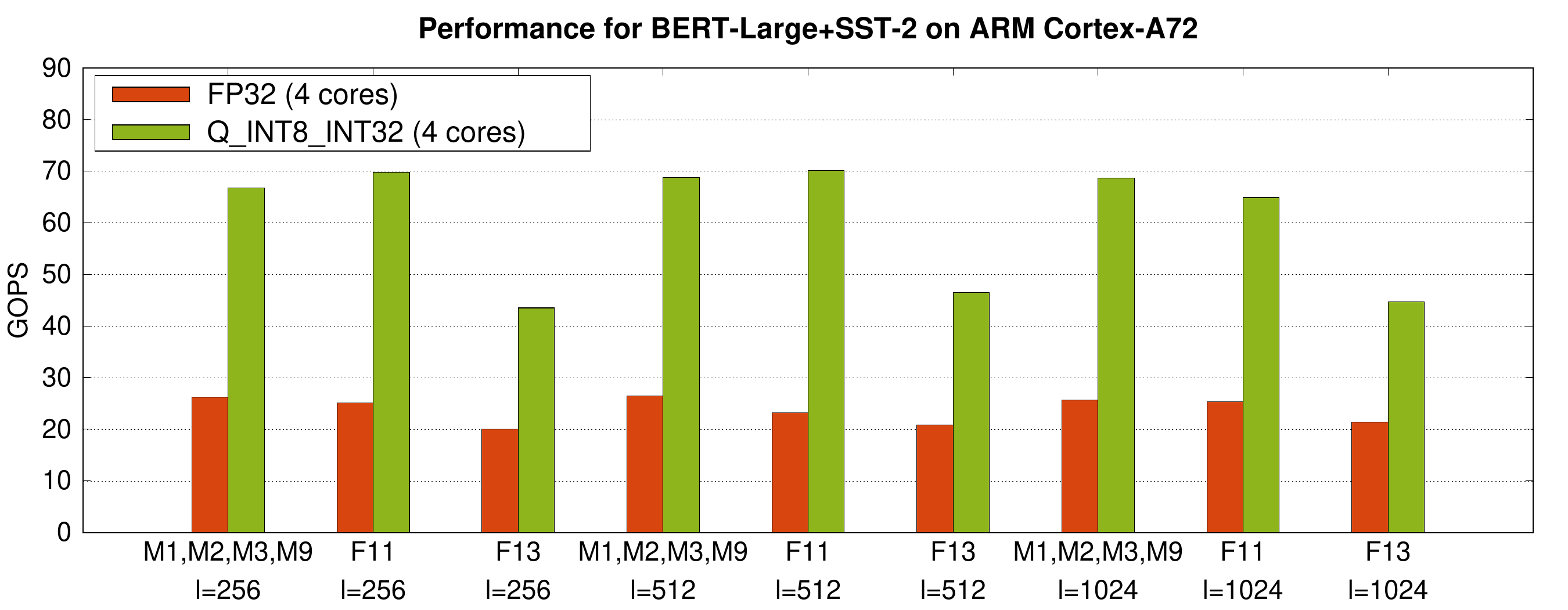} 
\includegraphics[width=0.98\columnwidth]{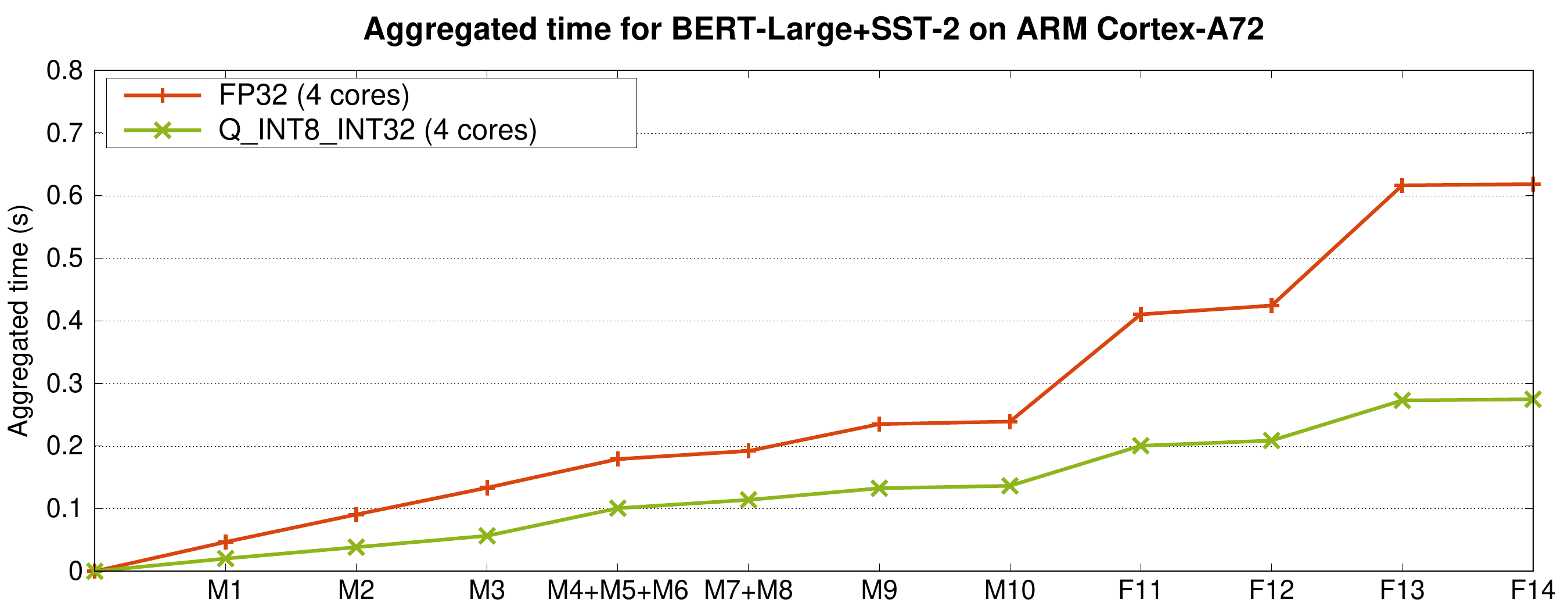} \\
\includegraphics[width=0.98\columnwidth]{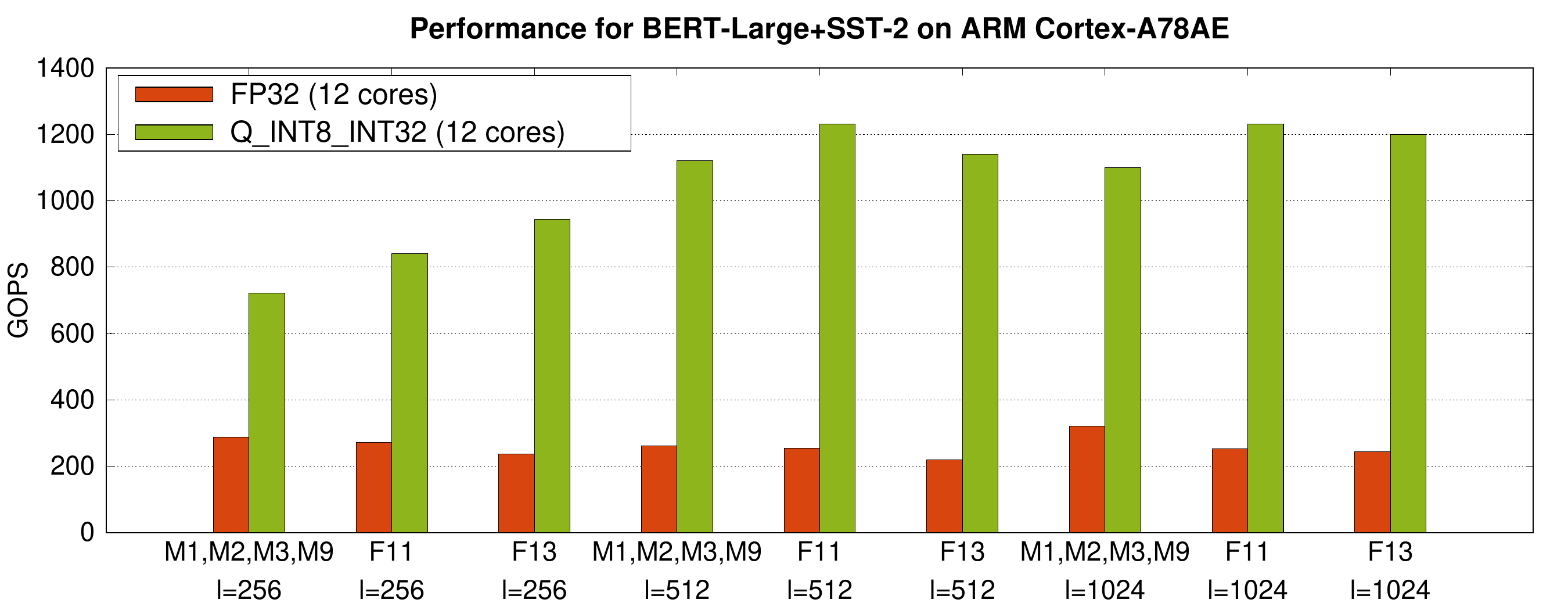} 
\includegraphics[width=0.98\columnwidth]{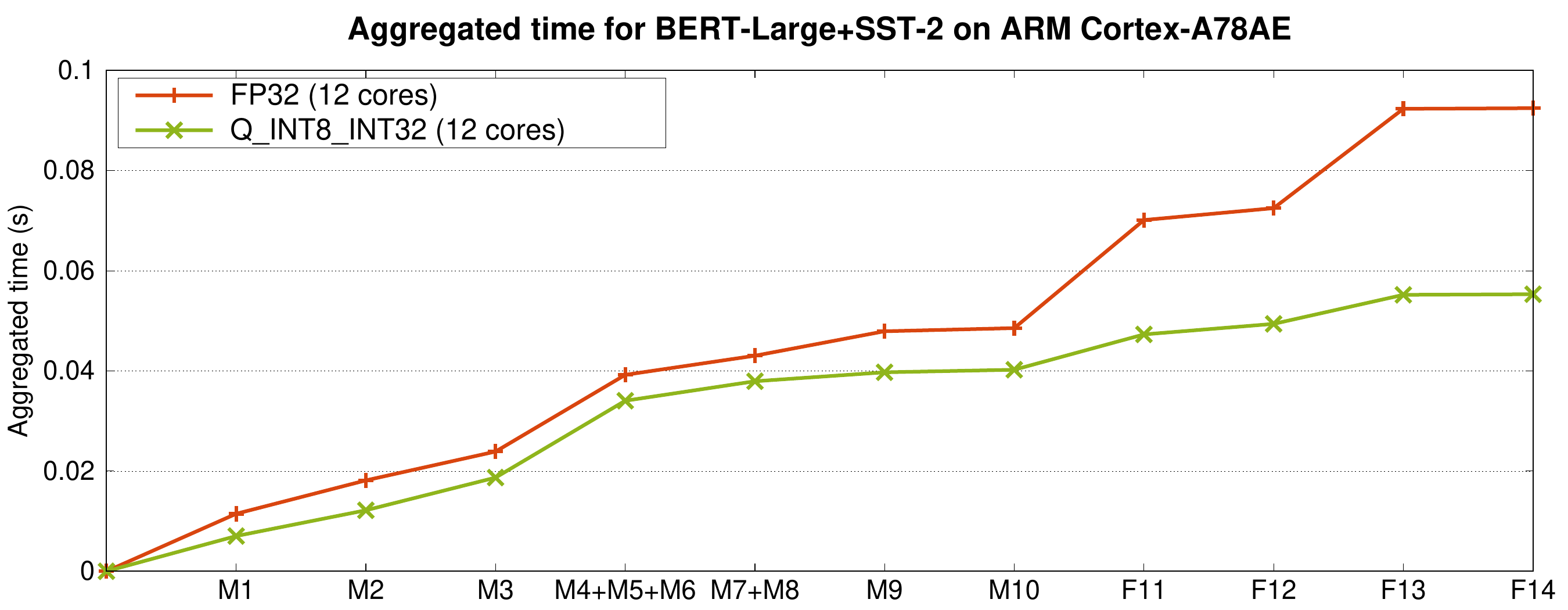} \\
\includegraphics[width=0.98\columnwidth]{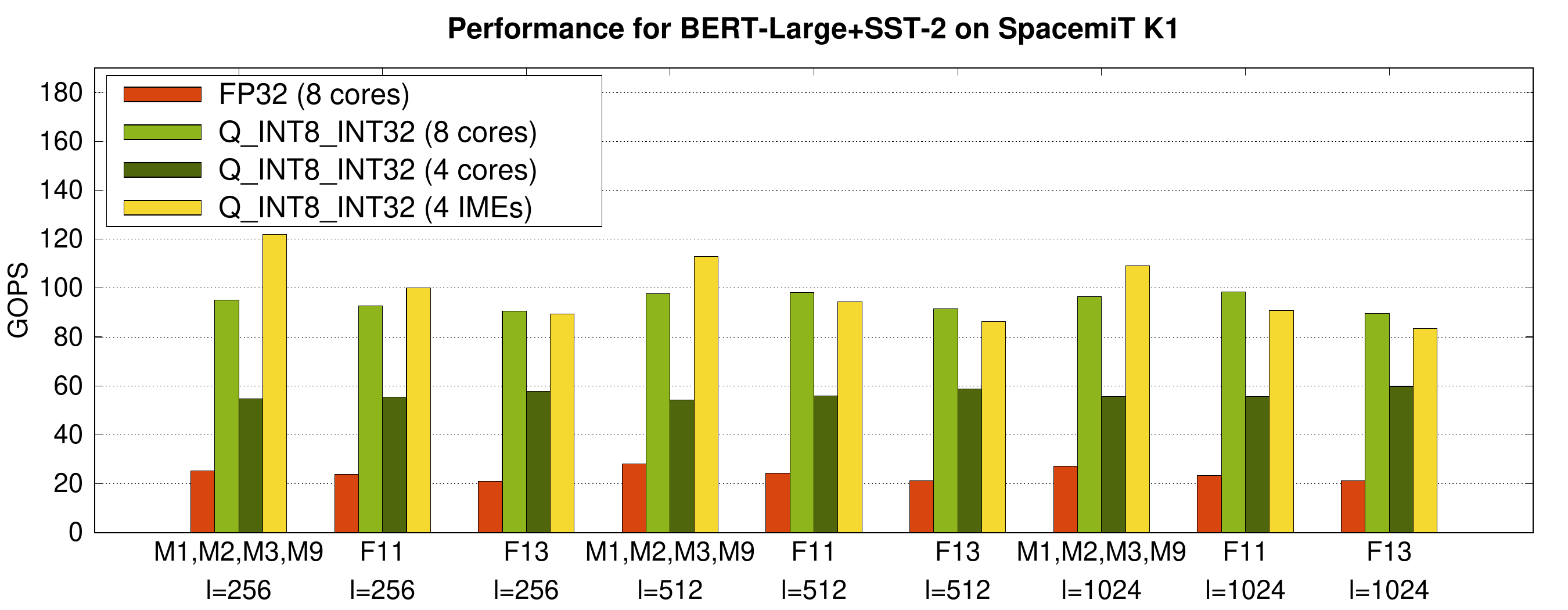} 
\includegraphics[width=0.98\columnwidth]{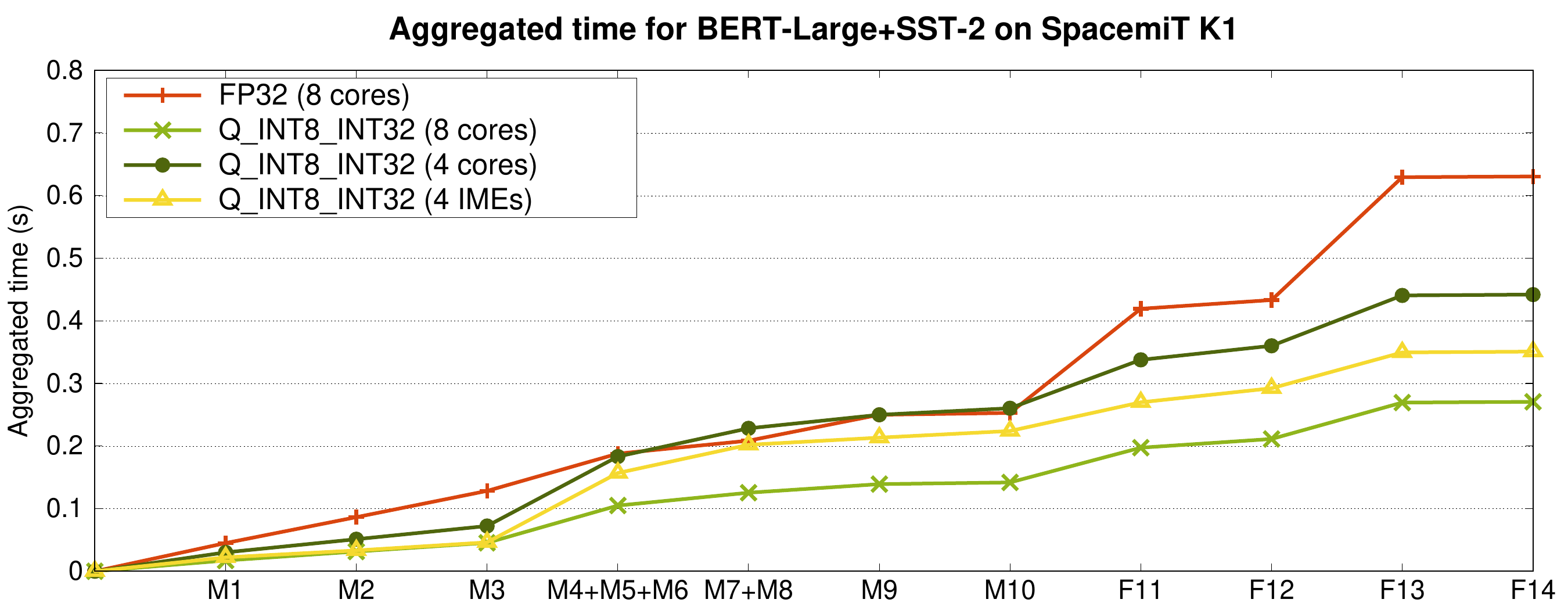} 
\caption{Performance of FP32 and MIP inference for a single encoder block 
         of BERT-Large model + SST-2 dataset. 
         Top row for ARM-Cortex A72, middle row for ARM-Cortex A78AE, and bottom row for SpacemiT K1. 
         Left column for individual layer performance (layers of the same dimension are grouped together), and right column for aggregated time (all layers). 
         The right-hand side plots
         correspond to the executions with $l=512$ tokens.}
\label{fig:bert_performance}
\end{figure*}

Figure~\ref{fig:bert_performance}
exposes the gains 
of MIP for the ``Large'' instance of the BERT family of
large language models (LLMs).
As in the previous case, the plots are organized into three rows, corresponding to the three platforms;
and two columns, for the arithmetic rates per operation and aggregated time.
Similar conclusions can be obtained from these experiments: 1) MIP outperforms FP32 in all the evaluated scenarios; 
2) the four IME accelerators in the SpacemiT K1 device deliver higher performance compared with the use of the same number of cores; 
3) however, their throughput is close but not higher than that observed when using the 8 cores of the entire SpacemiT K1 CPU. 
}

For the aggregated time (right-hand side plots in the same figure),
the speed-ups attained by the MIP solutions with respect to the reference FP32 model are 2.25$\times$, 1.67$\times$, and 2.32$\times$ for the 
ARM Cortex-A72, ARM Cortex-A78AE, and Space\-miT K1 with 8 cores, respectively.
\textcolor{black}{Regarding the SpacemiT K1 with 4 cores and 4 IMEs, the speed-ups are 1.42$\times$ and 1.79$\times$ respectively.}

{\color{black} Table~\ref{tab:bert_accuracy} reflects the minimal impact of the quantization process on the different evaluation metrics. As in the ResNet50v1.5 accuracy results, the maximum drop is around 1\%.}

\begin{table}[th!]
{\color{black} \small
\begin{center}

\begin{tabular}{lccccc}
\textbf{Model} & \textbf{Accuracy} & \textbf{F1} & \textbf{MCC} & \textbf{Precision} & \textbf{Recall} \\
\midrule
FP32 & 0.944 & 0.944 & 0.887 & 0.944 & 0.944 \\
Quantized & 0.932 & 0.931 & 0.863 & 0.932 & 0.932 \\
\midrule
\end{tabular}
\end{center}
\caption{\textcolor{black}{Accuracy of the original (FP32) and quantized  (INT8+INT32) BERT-Large models on SST-2.}}\label{tab:bert_accuracy}
}
\end{table}

\section{Concluding Remarks}
\label{sec:remarks}

Mixed-precision arithmetic units, increasingly incorporated into modern CPUs to enhance DL inference in edge computing contexts, offer the potential for substantial reductions in energy consumption, execution time, memory usage, and bandwidth demands. However, fully realizing these advantages necessitates meticulous software-level optimization to align with the capabilities and constraints of the target hardware.

In this work we have detailed
the implementation of the \gemm\ micro-kernel and its associated data layout, 
customized for the ISA of various platforms, including ARMv8.0-A NEON, ARMv8.2-A NEON, ARM SVE2, ARM SME, Intel AMX, and RISC-V with propietary matrix extensions.
In addition, we have showcased the performance benefits of MIP across inference workloads in computer vision and natural language processing on three representative platforms: an ARM Cortex-A72 CPU, ARM Cortex-A78AE CPU, and a RISC-V SpacemiT K1 CPU with dedicated integer multiplication accelerators.

For the evaluated DL benchmarks, our results show that quantized models deliver substantial improvements in execution time with minimal impact on accuracy. Although energy consumption was not fully  measured in this study, we also anticipate significant savings from using quantized models over FP32, given their reduced computational and memory demands. For example, 
the execution of a single-image inference
for Res\-Net50v1.9+ImageNet on the ARM Cortex-A78AE processor, required 0.98~Joules 
when using the original model 
but only 0.20~Joules with the quantized version (a gain of 5.1$\times$);
for 
BERT-Large+SST-2 and sequence length $l=512$, batch size $b=1$, 
the consumption of the original case and the quantized version
was 0.79 and 0.29~Joules (a gain of 2.72$\times$).
Finally, despite not all operands being stored in INT8, quantization reduced memory requirements to approximately 25\% of the original model in the case of ResNet50v1.5+ImageNet and batch size $b=1$;  
and to 35\% in the case of BERT-Large+SST-2 and $l=512$, $b=1$.

\section*{Acknowledgments}
This work is supported by grants
 PID2021-126576NB-I00 and PID2023-146569NB-C2
of MCIN/AEI/10.13039/5011000 11033, by {\em ``ERDF A way of making Europe''}, {\color{black}Programa PROMETEO 2023 (CIPROM/2022/20) by Conselleria de Innovación, Universidades, Ciencia y Sociedad
Digital from the Generalitat Valenciana,} and by Ayuda a Primeros Proyectos de Investigación (PAID-06-24), Vicerrectorado de Investigaci\'on de la Universitat Politècnica de València. 
A. Castell\'o is supported by the CIAPOS/2023/431 grant, and Fondo Social Europeo Plus 2021-2027 from the Generalitat Valenciana.

\bibliographystyle{elsarticle-num} 

\bibliography{biblio,deep,archs,codegen}
\balance

\end{document}